\documentclass{article}

\usepackage[preprint]{neurips_2024}

\usepackage[utf8]{inputenc} 
\usepackage[T1]{fontenc}    
\usepackage{hyperref}       
\usepackage{url}            
\usepackage{booktabs}       
\usepackage{amsfonts}       
\usepackage{nicefrac}       
\usepackage{microtype}      
\usepackage{xcolor,colortbl}         
\usepackage{amsmath}
\usepackage{amssymb}
\usepackage{bbm}
\usepackage{pifont}
\usepackage{multirow}
\usepackage[pdftex]{graphicx}
\usepackage{subcaption}
\usepackage{wrapfig}
\usepackage{fancyhdr}
\usepackage{amsthm}
\usepackage{xspace}
\usepackage{makecell}
\usepackage{capt-of}
\usepackage{sidecap}

\newcommand{\cmark}{\ding{51}}
\newcommand{\xmark}{\ding{55}}

\newcommand\ie{i.e.\xspace}
\newcommand\eg{e.g.\xspace}

\newcommand{\x}{\boldsymbol{x}}
\newcommand{\y}{\boldsymbol{y}}

\newcommand{\probpr}{\mathrm{Pr}}

\newcommand{\bv}{\boldsymbol{v}}

\newcommand{\bL}{\boldsymbol{L}}
\newcommand{\bP}{\boldsymbol{P}}
\newcommand{\btheta}{\boldsymbol{\theta}}
\newcommand{\bmu}{\boldsymbol{\mu}}
\newcommand{\sce}{\mathrm{e}}

\theoremstyle{plain}
\newtheorem{theorem}{Theorem}[section]

\theoremstyle{definition}
\newtheorem{definition}[theorem]{Definition}

\theoremstyle{remark}

\usepackage{amsmath,amsfonts,bm}

\def\1{\bm{1}}

\DeclareMathAlphabet{\mathsfit}{\encodingdefault}{\sfdefault}{m}{sl}
\SetMathAlphabet{\mathsfit}{bold}{\encodingdefault}{\sfdefault}{bx}{n}

\newcommand{\normltwo}{L^2}

\title{On Partial Prototype Collapse in the \\DINO Family of Self-Supervised Methods}

\author{%
Hariprasath Govindarajan$^{1,2}$ \quad Per Sidén$^{1,2}$ \quad Jacob Roll$^2$ \quad Fredrik Lindsten$^1$ \\
$^1$Linköping University, Sweden \quad $^2$ Arriver Software AB\\
\texttt{\{hargov,psiden,jroll\}@qti.qualcomm.com}, \\
\texttt{fredrik.lindsten@liu.se}\\
}

\begin{document}

\maketitle

\begin{abstract}
  A prominent self-supervised learning paradigm is to model the representations as clusters, or more generally as a mixture model. Learning to map the data samples to compact representations and fitting the mixture model simultaneously leads to the representation collapse problem. Regularizing the distribution of data points over the clusters is the prevalent strategy to avoid this issue. While this is sufficient to prevent full representation collapse, we show that a partial prototype collapse problem still exists in the DINO family of methods, that leads to significant redundancies in the prototypes. Such prototype redundancies serve as shortcuts for the method to achieve a marginal latent class distribution that matches the prescribed prior. We show that by encouraging the model to use diverse prototypes, the partial prototype collapse can be mitigated. Effective utilization of the prototypes enables the methods to learn more fine-grained clusters, encouraging more informative representations. 
  We demonstrate that this is especially beneficial when pre-training on a long-tailed fine-grained dataset. 
\end{abstract}

\section{Introduction}\label{sec:intro}

Self-supervised learning (SSL) is an effective approach to learn representations from unlabelled datasets. SSL methods have progressed rapidly in recent years and even surpassed the performance achieved by supervised training on several downstream tasks \cite{byol, moco_v3, dino, ibot, mae}. Broadly, SSL methods can be categorized into contrastive and non-contrastive methods. 
In contrastive methods \citep{ssl_simclr, moco, ssl_pirl}, all data samples repel all other data samples resulting in an approximately uniform distribution of representations in the latent space \citep{understanding_contrastive_ssl}. 
Specialized techniques like memory banks \citep{moco, ssl_pirl} have enabled such methods to perform well even without large batch sizes. 
Recent state-of-the-art SSL methods \citep{byol, simsiam, ibot, mae} use Vision Transformers \citep{vit} and non-contrastive training methods. The prototypical formulations used in the DINO family of methods \citep{dino, esvit, ibot, dino_vmf, dinov2} enable data samples belonging to the same semantic cluster to concentrate while only repelling other clusters. Such methods learn representations that are effective at nearest neighbor tasks and few-shot learning.

\begin{figure}[t]
    \centering
    \includegraphics[width=0.95\linewidth]{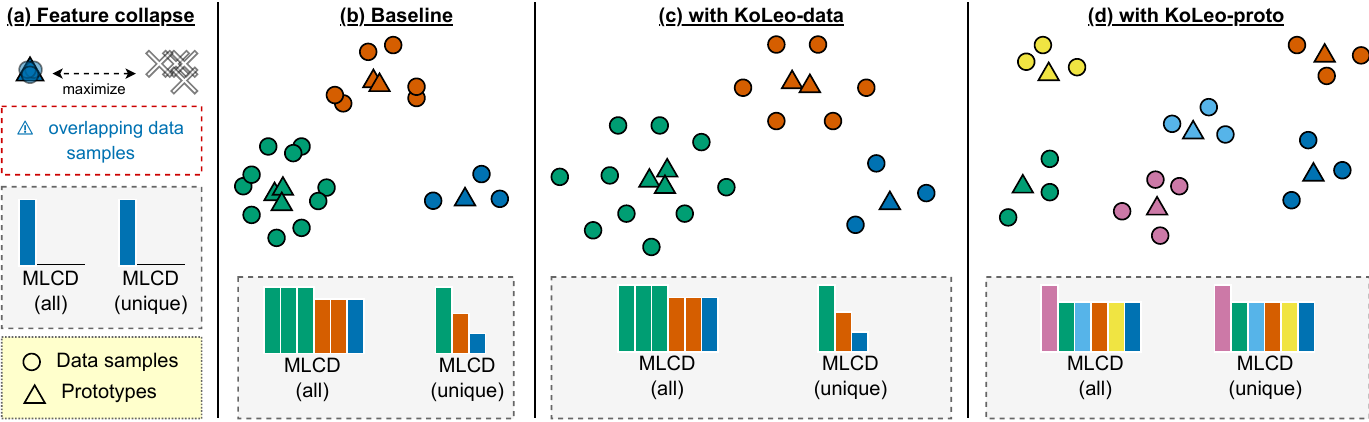}
    \caption{(a) The DINO family of methods result in a trivial full representation collapse without any regularization. (b) Using MLCD regularization such as centering and sharpening prevents full representation collapse but a partial prototype collapse still occurs. (c) KoLeo-data proposed in \citet{dinov2} spreads the data representations further apart but does not address the partial prototype collapse. Note that the method (both baseline and with KoLeo-data) uses the partial prototype collapse to achieve a MLCD closer to a uniform distribution over all the prototypes. But the MLCD over only the unique prototypes is non-uniform. (d) We propose KoLeo-proto regularization that explicitly encourages diverse prototypes and prevents partial prototype collapse.}
    \label{fig:intro_ppc_koleo}
\end{figure}

A common problem in this family of methods is the representation collapse. This originates from the simultaneous learning of the image representations as well as the clustering parameters. All existing methods regularize the marginal latent class distribution in order to prevent collapse. We show that these methods are still affected by a partial prototype collapse (\ie some groups of prototypes converge to the same vector), resulting in much fewer unique prototypes compared to the initialized number ($K$). We consider a prototype to be unique if it is at least $\epsilon$ distance away from all other prototypes. We show an illustration of this in Figure~\ref{fig:intro_ppc_koleo}. Moreover, varying the hyperparameter $K$ has limited effect on the number of unique prototypes. The consequence is that the number of learned clusters cannot be reliably controlled through the hyperparameter. Hence, it is thus far unclear what impact varying the number of clusters will have on these methods. 
In addition, we show that this shortcut is exploited by the methods to achieve a mean probability distribution that is closer to the prescribed distribution (usually a uniform distribution). 

\textbf{Contributions:} 
We formally define a partial prototype collapse and demonstrate its occurrence in the DINO family of methods, one of the most prominent family of SSL methods currently.
We propose KoLeo-proto regularization to prevent such a collapse by explicitly encouraging diverse prototypes by maximizing their differential entropy. Then, we study the downstream impact of effective utilization of the prototypes. 
For datasets like Imagenet with uniform class distribution, we find this to be beneficial for few-shot learning (FSL) and marginally improves performance in full data scenarios. However, we observe a trade-off that exists between FSL performance on the pre-training dataset and transfer performance, that is consistent with other methods that report improved FSL performance. 
When pre-training on a long-tailed dataset such as iNaturalist-2018, we observe a clear performance gain when classifying the same dataset without affecting the transfer performance. 

\section{Background}

The DINO-family of methods \citep{dino, ibot, msn, pmsn, esvit, dino_vmf} use the pretext task of assigning data to $K$ latent classes with multi-view class consistency. 
Consider an encoder model that produces a $\normltwo$-normalized representation $\y = g_{\btheta}(\x)$ such that $\Vert \y \Vert = 1$, for a data point $\x$ using parameters $\btheta$. The probability of assigning a data point to a latent class $k$ under the assumption of a latent class prior $\pi_k$ is given by: $P_k(\y) = \probpr(z=k | \y) = \frac{\pi_k \probpr(\y | z=k)}{\sum_{j=1}^K \pi_k \probpr(\y | z=j)}$.
With a uniform class prior $\pi_k \equiv 1/K$ (which is true in most prior work \citep{msn}),
\citet{dino_vmf} showed that the prototypical formulation in the DINO family corresponds to a von Mises-Fisher mixture model, with parameters $\{\bmu_k, \kappa_k\}$ and a normalization constant $C_p(\kappa_k)$  
\begin{equation}
\label{eq: dino_vmf_distribution}
    P_k(\y) = \frac{C_p(\kappa_k) \exp \langle \kappa_k \bmu_k, \y \rangle }{\sum_{j=1}^K C_p(\kappa_j) \exp \langle \kappa_j \bmu_j, \y \rangle } .
\end{equation}
Here, $\bmu_k$ is the mean vector (a.k.a prototype) with $\Vert \bmu_k \Vert = 1$ and $\kappa_k > 0$ is the precision, which is a measure of concentration around the mean vector.
The pre-training objective minimizes the KL-divergence between the latent class distributions of multiple views of each image. 
This task has a trivial solution where all data points can be mapped to the same representation. To prevent this collapse, it is essential to add some form of regularization to the training objective. The regularization techniques used in such methods can be motivated using the two requirements: (i) \textit{the model should learn distinct clusters} and (ii) \textit{spread the data over all these clusters}. The collapse where one or a few components dominate violates requirement-(ii). The collapse of individual probability distributions to uniform distributions implies that all the prototypes are equidistant from all the data representations. In practice, this leads to all prototypes collapsing to the same vector, which violates requirement-(i). 

\textbf{Connection between DINO and contrastive learning:} Contrastive learning typically uses the normalized temperature-scaled cross entropy loss based on cosine similarities. Then, the probability distribution of a query representation $\y_q$ being similar to a set of candidate representations $\y_k$ is defined as: $P_k(\y_q) = \frac{\exp(\langle \y_q, \y_k \rangle / \tau)}{\sum_{j=1}^K \exp(\langle \y_q, \y_j \rangle / \tau)}$.
SimCLR \citep{ssl_simclr} uses candidate representations from the same batch and MoCo \citep{moco} uses a memory bank instead to avoid large batch sizes. Comparing this to Eq.~\eqref{eq: dino_vmf_distribution}, one can observe that the prototypes in DINO can be viewed as exemplary representatives of the dataset, replacing the memory bank. Thus, the DINO family of methods are a sparse variant of sample-contrastive methods \citep{ssl_duality}.

\section{Marginal latent class distribution}
\label{sec: mlcd}

Before discussing a newly identified mode of collapse in the next section, we review and provide a unified understanding of some of the regularization techniques proposed in the literature to avoid collapse. 
We define the marginal latent class distribution (MLCD) as the probability vector with elements, $\Bar{p}_k = \mathbb{E}_{\x}[P_k(g_{\btheta}(\x))]$. To our knowledge, all existing methods avoid representation collapse by regularizing the MLCD. 
Specifically, the MLCD is encouraged to match a prescribed prior distribution. 
A uniform prior is the default choice except for \citet{pmsn}, who propose a power law distribution to better adapt the model to long-tailed data. 
In a self-distillation setup, the MLCD can be encouraged to match a prior distribution either by adjusting the teacher/target distributions or by adding a penalty on the online/student distributions. 

Adjusting the target distributions such that the MLCD matches a prior distribution can be posed as an entropy-regularized optimal transport problem, which can be solved using the Sinkhorn-Knopp (SK) algorithm \citep{sinkhorn_cuturi}. SK is typically run for a few iterations and adds a small but noticeable computational overhead. \citet{dino} proposed centering, a simpler and computationally efficient method to adjust the target distributions. A key distinction between Sinkhorn-Knopp and centering is that they adjust the target distributions $P^{\mathrm{(t)}}_k(\y)$ based on batch and moving average estimates of the MLCD, respectively. On the other hand, \citet{msn,pmsn} add a prior-matching penalty on the batch-estimates of MLCD obtained from the online distributions $P^{\mathrm{(o)}}_k(\y)$. The penalty is defined as the KL-divergence between the MLCD and the prior distribution. With a uniform prior, this is equivalent to maximizing the entropy of MLCD, known as mean entropy maximization. 

\textbf{Is the centering adjustment ad-hoc?} At first glance, the centering adjustment in DINO might appear somewhat ad-hoc. 
However, we find that probability centering (PC) \citep{dino_vmf} is closely connected to SK. Consider a batch of $B$ logit scores over $K$ latent classes $\bL \in \mathbb{R}^{B \times K}$ and corresponding probability distributions $\bP$. The SK adjusted (1 iteration) probability distributions are obtained as follows (derivation in \ref{appendix: sk_centering_comparison} of supplementary):
\begin{equation}
\label{eq:sinkhorn_knopp_adjustment}
    \Tilde{\bP}_{b, k}^{\mathrm{(sk1)}} = \frac{\exp (\bL_{b, k} - \log (\frac{1}{B} \sum_b \bP_{b,k}))}{\sum_{j=1}^K \exp (\bL_{b, j} - \log (\frac{1}{B} \sum_b \bP_{b,j}))}.
\end{equation}
On the other hand, the probability centered distributions 
are obtained as follows, where the centering parameter $c_k$ is calculated as a moving average estimate with momentum rate $m$:
\begin{align}
\label{eq:centering_adjustment}
\Tilde{\bP}_{b, k}^{\mathrm{(pc)}} = \frac{\exp (\bL_{b,k} - c_k)}{\sum_{j=1}^K \exp (\bL_{b,j} - c_j)} ; \;\;\;    c_k \leftarrow mc_k + (1-m) \log \left[ \frac{1}{B} \sum_{b=1}^B \bP_{b,k} \right].
\end{align}

Comparing Eq.~\eqref{eq:sinkhorn_knopp_adjustment} and Eq.~\eqref{eq:centering_adjustment}, we observe that probability centering is equivalent to one iteration of SK with the key distinction that the logit adjustment is calculated as a moving average instead of a batch estimate. We compare them empirically in \ref{sec: expt_mlcd_regularization} of supplementary.

\clearpage
\section{Partial prototype collapse}

\begin{wraptable}{r}{0.458\linewidth}
    \vspace{-0.2cm}
      \centering
      \scriptsize
      \begin{tabular}{llll}
        \toprule
        Backbone & Method & \makecell{Initialized\\prototypes\\($K$)} & \makecell{Unique\\prototypes\\($M$)} \\
        \midrule
            ViT-S/16 & DINO-vMF & 65536 & 1157 \\
            ViT-B/16 & DINO-vMF & 65536 & 939 \\
            ViT-B/16 & iBOT & 8192 & 875 \\
            ViT-B/16 & iBOT-vMF & 8192 & 1170 \\
            ViT-L/16 & iBOT & 8192 & 969 \\
            ViT-S/16 & MSN$^*$ & 8142 & 3363 \\
            ViT-S/16 & PMSN$^*$ & 8142 & 3005 \\
        \bottomrule
      \end{tabular}
  \captionof{table}{Number of unique prototypes in existing models with $\epsilon = 0.025$ (default pre-training: ImageNet-1K, $*$: iNat-2018)}
  \label{table:existing_models_num_unique_prototypes}
\end{wraptable}
Regularizing the MLCD enables the methods to meet the requirement of spreading data over
clusters. However, the MLCD depends on both the data representations and the prototypes. 
Given a set of frozen data representations, a method can achieve MLCD matching a prior distribution simply by manipulating the prototypes (see Figure~\ref{fig:intro_ppc_koleo}). 
This holds true for all existing methods in the DINO family, as they all regularize the MLCD in a similar manner, as discussed in the previous section.
Sharpening prevents the extreme case when all prototypes collapse to the same vector. However, except for this limited guardrail, the existing regularization techniques do not ensure that the methods learn unique prototypes. We define the term \textit{partial prototype collapse}, where only a significantly small proportion of the learned prototypes are unique. 

\begin{definition}[Partial prototype collapse]
Consider the set $W = \{\bmu_k : k=1,...,K\}$ of $K$ prototype vectors, $\bmu_k$ such that $\Vert \bmu_k \Vert = 1$. A \emph{partial prototype collapse} (of degree $M$ and $\epsilon$ distance) is said to have occurred if there exists a set of $M$ disjoint partitions of prototype vectors $V_m \subset W$, $m = 1, ..., M$, and $M$ representative prototype vectors $\bv_m \in V_m$, such that for all $m = 1, ..., M$, 
$1-\bv_m^{\mathsf{T}}\bmu_j < \epsilon$, for all $\bmu_j \in V_m$.
The set of $M$ \emph{unique prototypes} is defined as $U = \{ \bv_m \}_{m=1}^M$. For each representative prototype, the \emph{redundancy factor} $r_m$ is defined as the size of the corresponding set partition, $r_m = |V_m|$.
\end{definition}

\begin{SCfigure}
    \includegraphics[width=.45\linewidth]{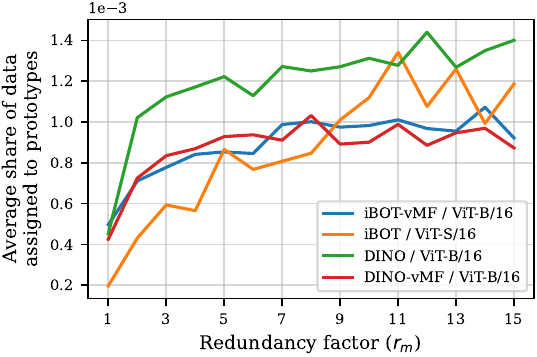}
    \caption{We reassign data to only the $M$ unique representative prototypes and compute the average proportion of data assigned to prototypes having specific redundancy factors. We find that the models tend to assign a larger proportion of data to prototypes with higher redundancy factors. This holds true for DINO and iBOT with different backbones.}
    \label{fig:avg_prob_redundant_components} 
\end{SCfigure}
\textbf{Investigating learned MLCD and prototypes}: 
When training with MLCD regularization, the DINO family of methods are prone to partial prototype collapse since it enables the method to spread probability mass associated with each unique prototype across its $\epsilon$-set of redundant prototypes. This acts as a shortcut to match the MLCD to the specified prior distribution. \citet{dino_vmf} make an empirical observation that the DINO models used significantly smaller number of unique prototypes compared to the hyperparameter $K$. 
However, this problem is neither studied further nor addressed by their proposed method. 
Based on our definition of partial prototype collapse and using a cosine distance metric, we investigate the prototypes learned by several self-supervised clustering methods that use a prototypical formulation, from SwAV \citep{swav} to iBOT \citep{ibot}. In Table~\ref{table:existing_models_num_unique_prototypes} and Table~\ref{table:existing_models_num_unique_prototypes_extended} (in supplementary), we show that such a collapse exists in all the considered methods. 
We observe that prototypes with a higher redundancy factor tend to be assigned a larger proportion of the data samples (see Figure~\ref{fig:avg_prob_redundant_components}). 
Hence, the partial prototype collapse serves as a shortcut to achieve a MLCD closer to the specified uniform prior in these works. This shortcut is important to be aware of, if the intention is to encourage the MLCD to match a specific non-uniform distribution based on knowledge about the dataset domain.  
In addition, this means that the hyperparameter $K$ does not play its intended role of controlling the number of clusters. 

\subsection{Regularizing prototype distribution}

The number of latent classes is an important choice in clustering as this controls the fine-grainedness of the clusters. Firstly, this controls the difficulty of the self-supervision task. Secondly, more informative representations are required to discriminate between more fine-grained latent classes. With this motivation, we believe that the number of prototypes is an important design choice in SSL. However, prior works have found inconsistent results when ablating for this choice, likely because of the occurrence of partial prototype collapse. 
Given that we want the prototypes to be as diverse as possible, a meaningful choice is to encourage the prototypes $\boldsymbol{W}=\{\bmu\}_{k=1}^K$ to be uniformly distributed in the latent space. We propose to achieve this by maximizing the differential entropy of the prototype vectors, obtained using the Kozachenko-Leonenko estimator \citep{koleo_estimate, koleo_estimator, spreading_vectors}, $h_{\mathrm{kl}}(\boldsymbol{W}) = - \frac{1}{K} \sum_{k=1}^K \log (d_k)$, where $ d_k = \min_{i \neq k} \Vert \boldsymbol{\mu}_k - \boldsymbol{\mu}_i \Vert $.
We efficiently compute an estimate of $\mathcal{L}_{\mathrm{KP}} = h_{\mathrm{kl}}(\boldsymbol{W})$ by randomly partitioning the prototypes into batches and we show in \ref{appendix: koleo_prototypes_implementation} of supplementary that this adds negligible computational overhead. We verify in section~\ref{sec: expt_prototype_regularization} that this regularization can mitigate the partial prototype collapse. Then, we focus on our main goal of studying the downstream impact of effectively utilizing the initialized prototypes through various experiments that evaluate the learned representations.

\section{Related Work}

\textbf{Connection to DINOv2}: Our proposed KoLeo-proto regularization is formulated similar to \citet{spreading_vectors}. Recently, DINOv2 \citep{dinov2} proposed the KoLeo-data regularization which uses a similar formulation but applied to spread the data representations instead of the prototypes. 
Hence, DINOv2 can be viewed as an interpolation between the uniformly distributed representations of contrastive learning and clustered representations of the DINO family. 
In contrast, KoLeo-proto preserves the clustered representations of DINO and encourages the method to learn diverse clusters. We illustrate this difference in Figure~\ref{fig:intro_ppc_koleo}.

\textbf{Regularizations in clustering-based SSL}: We provide an extended discussion of clustering based SSL methods in \ref{appendix: extended_related_work} of supplementary and focus our discussion on the regularization methods in this section. 
A common limitation of simultaneously learning representations and clustering them is that there are degenerate solutions that perfectly solve the clustering task but fail to learn informative representations. 
\citet{deep_cluster} proposed uniform pseudo-label sampling that is equivalent to weighting the loss contribution of an input by the inverse of its assigned cluster's size. 
\citet{sela} and \citet{swav} viewed the clustering task with MLCD regularization as an entropy regularized optimal transport problem and use the Sinkhorn-Knopp (SK) algorithm to assign pseudo-labels to data points \citep{sinkhorn_cuturi}. 
\citet{sela} and \citet{swav} used the Sinkhorn-Knopp (SK) algorithm to regularize the MLCD \citep{sinkhorn_cuturi}. 
This is shown by \citet{pmsn} to encourage the MLCD to match a uniform prior. While SK requires multiple iterations for convergence, a simpler and computationally cheaper approach known as centering is proposed in DINO \citep{dino} and also used in EsViT \citep{esvit} and iBOT \citep{ibot}. \citet{dino_vmf} proposed probability centering, that computed the centering parameter in the probability space instead of the logit space. MSN \citep{msn} and PMSN \citep{pmsn} proposed to add an explicit prior matching penalty to encourage the MLCD to align with prescribed prior distributions. Methods using the prior-matching penalty and SK depend on batch estimates of the MLCD. On the other hand, centering uses moving average estimates; we showed the connection of probability centering to SK in section~\ref{sec: mlcd}.
and investigate how methods using batch and moving average MLCD estimates compare at different batch sizes in section~\ref{sec: expt_mlcd_regularization}. 
While all the above methods regularize the MLCD, we show the occurrence of a partial prototype collapse by investigating the prototypes learned by existing pre-trained models. We propose a new KoLeo-proto regularization as a tool to prevent this collapse and study the downstream impact of effectively utilizing the prototypes. 

\textbf{Pre-training on long-tail datasets}: 
Most SSL methods are evaluated by pre-training on ImageNet, a well-curated dataset with a uniform class distribution. To the contrary, real-world data collection often results in long-tailed distributions over visual concepts and pre-training on such datasets is of practical interest. We note that there is limited research on pre-training SSL methods on such long-tailed datasets. 
\citet{ssl_deepercluster} investigated pre-training on a large uncurated dataset. Recently, \citet{temperature_schedules} explored the benefits of using temperature schedules in the context of contrastive learning. 
A custom temperature schedule is already used in the DINO family of methods, and found to perform better than constant temperatures. 
\citet{pmsn} showed that pre-training on a long-tailed dataset can benefit from choosing an appropriate long-tail prior. We investigate the impact of effective prototype utilization when pre-training on a long-tailed dataset in section~\ref{sec: inat18_experiments}. 
\citet{etf_neural_collapse} overcame a minority collapse issue \citep{minority_collapse} in supervised long-tailed classification with a fixed classification layer based on equiangular tight frames (ETF) geometry. However, this comes with the implicit assumption that all class prototypes should be equidistant which is a strong assumption for the latent classes learned in SSL. 
The KoLeo-proto regularization is a soft penalty that encourages the prototypes to remain distinct while still allowing semantically similar clusters to have more similar prototypes. This enables the SSL method to learn better semantically meaningful representations.

\section{Experiments}
\label{sec: imagenet_experiments}

To study the MLCD and prototype regularizations, we focus on iBOT, which is a strong recent baseline among the DINO family of methods and also used as the foundation for DINOv2 \citep{dinov2}. 
We pre-train the models on the ImageNet-1K dataset \citep{dataset_imagenet} by modifying the public codebase of iBOT \footnote{https://github.com/bytedance/ibot}. 
We use the same hyperparameter settings as in iBOT for different ViT backbones 
(see configuration details in \ref{appendix: hyperparameter_settings} and additional experiment with a CNN backbone in \ref{appendix: imagenet_dino_resnet_expt} of supplementary) and use the vMF normalized variants \citep{dino_vmf}, which are shown to produce stable trainings and improved performance. 
We start with ablation experiments to choose the MLCD regularization technique in section~\ref{sec: expt_mlcd_regularization}, evaluate the impact of adding our proposed prototype regularization in section~\ref{sec: expt_prototype_regularization} and finally perform full-scale pre-training experiments.

\subsection{MLCD regularization}
\label{sec: expt_mlcd_regularization}

\begin{wrapfigure}{r}{0.4\textwidth}
  \vspace*{0.0cm}
  \centering
    \includegraphics[width=\linewidth]{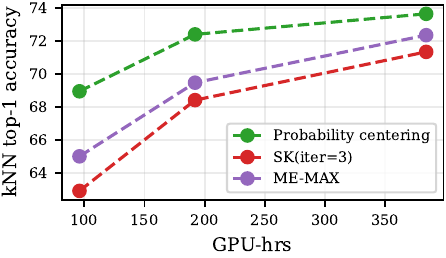}
    \caption{ImageNet top-1 kNN accuracy with different MLCD regularizations. Probability centering performs better than SK and ME-MAX at different compute budgets.}
    \label{fig:ablation_mlcd_regularization}
\end{wrapfigure}
We conduct ablation experiments to select the method to regularize MLCD. We pre-train ViT-S/16 backbone with different MLCD regularization techniques - Sinkhorn-Knopp (SK), probability centering (PC) and mean entropy maximization (ME-MAX). For PC, we use the vMF normalized version of iBOT. For SK and ME-MAX, we chose to use a smaller teacher temperature based on a hyperparameter search (refer \ref{appendix: hyperparameter_settings} for details). We also consider three different compute budgets (2, 4 and 8 GPUs for 2 days), which allows us to evaluate the impact of batch size on these techniques. With more GPUs, we can accommodate a larger batch size. The number of epochs is adjusted such that the total number of iterations are the same for all the compute budgets. We do this to avoid the expensive process of optimizing the learning rates for each compute budget and regularization method. Overall, from Figure~\ref{fig:ablation_mlcd_regularization}, we find that probability centering performs better than the other alternatives at different compute budgets. Interestingly, PC achieves performance on par or better than the alternatives, even at half of the compute budget (\eg PC/4GPUs vs ME-MAX/8GPUs). 

The methods discussed above have all been proposed in the literature as ways to regularize the MLCD. We argue that the main difference between them is whether the regularization is done over a single batch (SK, ME-MAX) or based on moving average statistics (PC). We observe that PC performs significantly better than the alternatives at the lowest compute budget, which uses a small batch size. As we increase the compute budget and thereby also the batch size, the gap is reduced. This indicates that PC is more robust to the choice of batch size. We conjecture that this is due to too noisy estimates of the MLCD when computed over a batch, which is not surprising considering that we estimate probability vectors in a high-dimensional space. 
Therefore, in all the experiments in the main paper, we use the vMF normalized iBOT with MLCD regularized using probability centering.

\subsection{Prototype regularization}
\label{sec: expt_prototype_regularization}

\begin{figure*}[t]
    \centering
    \begin{subfigure}{.45\linewidth}
        \centering\includegraphics[width=\linewidth]{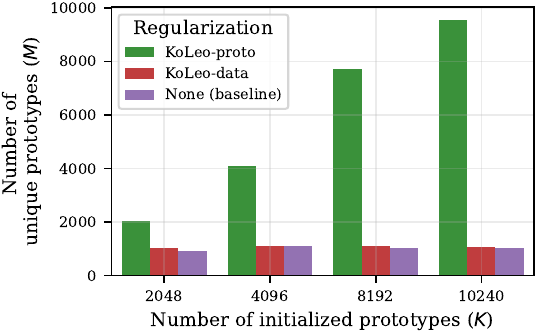}
        \label{fig:koleo_ablation_num_unique_protos}
    \end{subfigure}
    \hspace{0.04\linewidth}
    \begin{subfigure}{.45\linewidth}
        \centering
        \includegraphics[width=\linewidth]{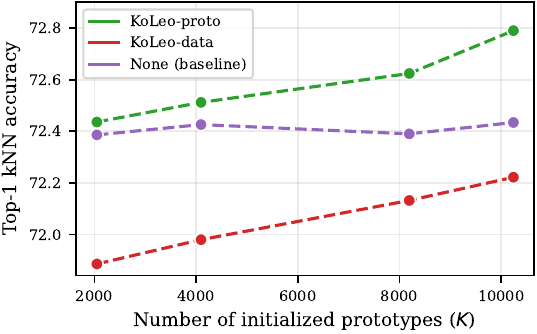}
        \label{fig:koleo_ablation_accuracies}
    \end{subfigure}
    \caption{(left) The number of unique prototypes are similar for the baseline and KoLeo-data regularization at different number of initialized prototypes. With KoLeo-proto, most of the initialized prototypes remain unique. This means that the hyperparameter $K$ can meaningfully control the number of learned clusters. (right) The number of initialized prototypes has no impact on the baseline performance. With any form of KoLeo-regularization, more prototypes lead to better performance and KoLeo-proto consistently performs best.}
    \label{fig:koleo_ablation_accuracy}
\end{figure*}

We add our proposed KoLeo-proto regularization to the iBOT-vMF baseline, resulting in the overall loss objective, $\mathcal{L} = \mathcal{L}_{\mathrm{iBOT}} + \lambda \mathcal{L}_{\mathrm{KP}}$. These results are indicated by "(kp)". Similarly, we indicate the KoLeo-data regularization used by \citet{dinov2} as "(kd)". 
We use $\lambda=0.1$ and observe that such a small $\lambda$ is sufficient to mitigate partial prototype collapse (see ablation in \ref{appendix: ablation_koleo_regularization} of supplementary). In Figure~\ref{fig:koleo_ablation_accuracy}, we compare the number of unique prototypes $M$ when we vary the initialized number of prototypes hyperparameter $K$. With the baseline and KoLeo-data regularization, changing $K$ has no impact on the number of unique prototypes learned by the method, which is significantly smaller than the initialized number of prototypes. This indicates the occurrence of partial prototype collapse. 
With KoLeo-proto regularization, we observe that $M \approx K$ and hence the hyperparameter $K$ reliably controls the number of learned clusters.

We observe that the baseline shows similar performance at different numbers of initialized prototypes. On the other hand, with KoLeo-data, the performance is worse than the baseline but continues to improve as the number of prototypes are increased. KoLeo-data encourages the data to spread on the hypersphere. Hence, data is assigned to more diverse prototypes compared to the baseline in the initial training phase. We conjecture that this initial training dynamic benefits from having more prototypes, even if many of these prototypes eventually collapse to the same vector. 
We limit the maximum number of prototypes to 10240 due to computational limitations. Computing probability distributions for all the tokens over more dimensions adds a large computational overhead. However, the KoLeo regularization itself only adds a negligible computational overhead (cf. \ref{appendix: koleo_prototypes_implementation} in supplementary). 
With KoLeo-proto, we observe around 0.1\% improvement in accuracy when adding every 2K additional prototypes. 
Overall, increasing the number of prototypes from 2K to 10K results in a 0.4\% improvement.  
Further scaling of the number of prototypes can bring larger performance gains which should be feasible with the efficient implementation in DINOv2. 
\begin{table*}[t]
  \centering
  \scriptsize
  \begin{tabular}{lllllllll}
    \toprule
    Method & kNN & Linear & Finetuning & 1\% data & 5 img/cls & 2 img/cls & 1 img/cls & Avg. Transfer \\
    \midrule
    \multicolumn{8}{l}{\textit{ViT-Base/16}}\\
    DINO-vMF & 77.4 & 78.8 & 83.6 & 70.4 & 66.1 & 59.3 & 50.3 & 86.8 \\
    MSN & 73.3 & 74.8 & -- & 69.1 & 65.5 & 58.9 & 49.8 & 85.3 \\
    WE-SSL & 77.2 & 78.9 & -- & 71.5 & \underline{68.3} & \textbf{62.4} & \textbf{53.7} & -- \\
    iBOT-vMF & \underline{78.7} & \underline{80.3} & \textbf{84.1} & \underline{72.3} & \underline{68.3} & 61.1 & 51.6 & \textbf{87.3} \\
    \rowcolor{white!60!lightgray} iBOT-vMF (kp) & \textbf{78.8} & \textbf{80.5} & \textbf{84.1} & \textbf{72.7} & \textbf{69.1} & \underline{62.0} & \underline{52.5} & \underline{86.6} \\
    \midrule
    \multicolumn{8}{l}{\textit{ViT-Small/16}}\\
    DINO-vMF & 74.7 & 77.0 & 81.8 & 65.0 & 59.1 & 49.4 & 39.2 & 85.4 \\
    MSN & 74.9 & 76.6 & -- & \underline{67.2} & \underline{62.8} & \underline{55.8} & \underline{47.1} & 84.1 \\
    WE-SSL & 75.2 & 77.4 & -- & \textbf{68.7} & \textbf{65.1} & \textbf{58.9} & \textbf{50.1} & 84.6 \\
    iBOT-vMF & \underline{75.3} & \textbf{77.9} & \textbf{82.3} & 66.4 & 60.6 & 51.1 & 40.7 & \underline{85.5} \\
    \rowcolor{white!60!lightgray} iBOT-vMF (kp) & \textbf{75.5} & \textbf{77.9} & \textbf{82.3} & 67.0 & 61.1 & 51.7 & 41.6 & \textbf{85.6} \\
    \bottomrule
  \end{tabular}
  \caption{ImageNet classification with full data and few-shot scenarios and transfer learning.}
  \label{table:imagenet_pretrained_classification_main}
\end{table*}

\subsection{Pre-training with ImageNet}

We pre-train iBOT-vMF with efficient prototype utilization using KoLeo-proto regularization for ViT-S/16 and ViT-B/16 backbones. To ensure fair comparison, we set the number of prototypes to 8192, similar to iBOT. Hence, any changes in performance can be associated to only the effective prototype utilization. In Table~\ref{table:imagenet_pretrained_classification_main}, we report the top-1 accuracies obtained using kNN and linear classification based on frozen backbone features, few-shot accuracies averaged over 3 different splits and the accuracy obtained after fine-tuning. For kNN, linear evaluation and finetuning, we follow the same protocol as in DINO and iBOT. We perform few-shot evaluation similar to \citet{msn} and use the provided data splits. We compare against the iBOT-vMF baseline, MSN and the best performing models from WE-SSL \citep{we_ssl}. We observe on par or marginal improvements for kNN, linear and fine-tuned classification performance. The kNN performance improvement with respect to the baseline at 8192 prototypes after full-scale pre-training mirrors the improvement (+0.2\%) observed after the small-scale ablation in Figure~\ref{fig:koleo_ablation_accuracy}. This suggests that by using an even larger number of prototypes one can improve the performance further with efficient prototype utilization (cf.  Figure~\ref{fig:koleo_ablation_accuracy}), which we found to not be the case for the baseline in ablation experiments. 

We find larger gains for few-shot learning (FSL) performance when adding KoLeo-proto to the baseline, even at 8192 prototypes. Note that the prediction head architecture and other hyperparameters are tuned in WE-SSL to achieve the best FSL performance with ViT-S/16. This explains the significantly better results achieved by WE-SSL with ViT-S/16. With ViT-B/16, iBOT-vMF (kp) outperforms WE-SSL at 1\% and 5 img/cls settings. Note that iBOT-vMF can be tuned similar to WE-SSL but we have not investigated this. Instead, we focus on studying the impact of effective prototype utilization on general downstream performance and do not perform any task-specific tuning.

\textbf{Transfer learning:} We conduct linear classification experiments on the standard suite of datasets trained using features extracted from a frozen pre-trained model. In Table~\ref{table:imagenet_pretrained_classification_main}, we report the accuracies averaged over all datasets. The detailed results and evaluation setup are provided in \ref{appendix: detailed_transfer_learning} of supplementary. 
We observe on par or decreased transfer performance with effective prototype utilization compared to the iBOT-vMF baseline. Interestingly, we note that the transfer performance decreases also in other methods that improve few-shot learning performance such as MSN \citep{msn} and WE-SSL \citep{we_ssl} compared to their DINO baseline (note that transfer results for ViT-B/16 are not reported in WE-SSL). 
This indicates that tuning for few-shot learning performance can potentially harm transfer performance. We hypothesize that certain features that improve few-shot learning performance on the pre-training dataset could be too specific to the pre-training data and do not generalize to other datasets considered for transfer learning.
Compared to these methods, our proposed regularization leads to better transfer performance. There appears to be a trade-off between few-shot learning performance on the pre-training dataset and transfer learning performance. Currently, it is unclear why such a trade-off exists and this requires further investigation. 
Note that DINOv2 constructs the LVD142M pre-training dataset by finding images similar to the suite of transfer datasets of interest, thus limiting the domain gap between the pre-training and transfer datasets.

\subsection{Pre-training with iNaturalist-2018}
\label{sec: inat18_experiments}

Most SSL methods are pre-trained on ImageNet which is well-curated and contains uniformly distributed data across its classes.
It is of practical interest to pre-train SSL methods on data collected in the wild, which is often long-tailed. 
We study pre-training the DINO family of methods on long-tailed datasets, which has gained limited attention. We consider the iNaturalist-2018 (iNat18) dataset \footnote{This dataset was used for academic purposes only.}
which is around 1/3rd of the size of ImageNet and contains a long-tail distribution of data from 8142 classes. We pre-train all the models for 300 epochs using the default publicly available hyperparameters. 
For MSN and PMSN \citep{msn, pmsn}, we choose the regularization strength $\lambda$ based on a hyperparameter search (see \ref{appendix: msn_pmsn_choice} in supplementary for details). 
This analysis of regularization strength $\lambda$ indicated that weakly encouraging a uniform prior in MSN produces better performance than using a long-tailed prior as in PMSN. 
With this motivation, we retain the uniform prior assumption of the other methods. In Table~\ref{table:inat2018_pretrained_classification}, we report the top-1 classification accuracy obtained using a linear and a fine-tuned classifier. 
For linear classification, we follow a similar protocol as in the ImageNet experiments. 
For fine-tuning, we use longer trainings with a smaller learning rate as in DINO \citep{dino} (see details in \ref{appendix: finetuning_recipes} of supplementary).
We consider iBOT-vMF as our baseline method, which significantly outperforms MSN and PMSN.

\begin{wraptable}{r}{0.57\linewidth}
  \vspace*{-0.2cm}
  \centering
  \scriptsize
  \begin{tabular}{lllll}
    \toprule
    Method & $M$ & Linear & Fine-tuned & Avg. Transfer \\
    \midrule
    \multicolumn{4}{l}{\textit{ViT-Small/16}}\\
    DINO-vMF & 1380 & 49.7 & 68.5 & 71.7 \\
    iBOT-vMF & 1804 & 50.1 & \textbf{69.4} & \underline{71.8}  \\
    iBOT-vMF (kd) & 1843 & \underline{50.5} & 69.1 & 71.5 \\
    \rowcolor{white!60!lightgray} iBOT-vMF (kp) & 7895 & \textbf{51.1} & \underline{69.3} & \textbf{72.0} \\
    MSN ($\lambda = 1$) & 3363 & 43.8 & 63.5 & 67.0 \\
    PMSN ($\lambda = 5$) & 3005 & 41.8 & 64.2 & 66.1 \\
    \midrule
    \multicolumn{4}{l}{\textit{ViT-Base/16}}\\
    iBOT-vMF (kd) & 1634 & 50.4 & 73.3 & 73.2 \\
    \rowcolor{white!60!lightgray} iBOT-vMF (kp) & 7573 & \textbf{51.4} & \textbf{74.0} & \textbf{73.7} \\
    \bottomrule
  \end{tabular}
  \captionof{table}{iNat-2018 (linear probing and fine-tuning) and avg. transfer classification accuracies}
  \label{table:inat2018_pretrained_classification}
\end{wraptable}
For ViT-S model, we find that both KoLeo regularization methods bring performance benefits compared to the baseline. After evaluating the two forms of KoLeo regularization on the ViT-B model as well, we conclude that KoLeo-proto regularization performs best. 
With partial prototype collapse, models learn more coarse-grained latent classes where the number of unique prototypes are less than the number of classes (see $M$ in Table~\ref{table:inat2018_pretrained_classification}). Then, the learned clusters are likely to have merged several of the fine-grained classes. This is mitigated when the prototypes are effectively utilized, leading to more diverse clusters and hence, more 
informative representations which are beneficial for long-tailed and fine-grained classification. Hence, iNat-2018 pre-training benefits more from effectively utilizing prototypes compared to the Imagenet experiments. 
We report average transfer learning accuracies 
in Table~\ref{table:inat2018_pretrained_classification} and detailed results in Table~\ref{table:linear_transfer_classification_inat18} of supplementary. 
In contrast to the ImageNet experiments, effective utilization of prototypes in KoLeo-proto is also beneficial for transfer performance.

\section{Conclusion}

We identified the occurrence of a previously unnoticed mode of collapse in the DINO family of methods, termed as \textit{partial prototype collapse} that results in significant redundancies in the prototypes. As a consequence, the hyperparameter controlling the number of prototypes did not perform its intended role of controlling the number of clusters learned by the model. We proposed the KoLeo-proto regularization to encourage the model to learn diverse prototypes. By adding our proposed regularization, we showed that the initialized prototypes are effectively utilized. With effective prototype utilization, scaling the number of prototypes is useful in learning better image representations of the underlying dataset. Using the same moderate number of 8K prototypes as before, we showed that few-shot learning performance can be improved and full data trainings can be marginally improved. As indicated in our ablation experiments, it seems possible that further scaling the number of prototypes can result in more significant improvements. However, we observed a worse transfer performance and this trade-off is consistent with other methods that specifically improve few-shot learning. 
On the other hand, we found that learning fine-grained clusters on a long-tailed fine-grained dataset such as iNat-2018 is more beneficial, indicated by the larger performance gains achieved using a similar number of prototypes. 

We have shown that the hyperparameter for the number of prototypes can be reliably controlled using our regularization. This has broad implications on applying methods from the DINO family. One can better understand the impact of using different numbers of clusters in the self-supervised pretext task for their own dataset and method of choice. This could vary depending on the domain of the dataset and how fine-grained the semantic concepts are in that domain. Computing probability distributions over a large number of latent classes comes at a significant computational cost (see \ref{appendix: computational_analysis} in supplementary). If indeed a small number of clusters are sufficient for some dataset, effectively utilizing fewer prototypes can help in reducing computational expenses.

\nocite{ssl_duality}

\begin{ack}
This research is financially supported by the Swedish Research Council via the project \emph{Handling Uncertainty in Machine Learning Systems} (contract number: 2020-04122),
the Wallenberg AI, Autonomous Systems and Software Program (WASP) funded by the Knut and Alice Wallenberg Foundation,
and
the Excellence Center at Linköping--Lund in Information Technology (ELLIIT). 
The computations were enabled by the Berzelius resource provided by the Knut and Alice Wallenberg Foundation at the National Supercomputer Centre. 
\end{ack}

\bibliographystyle{unsrtnat}
\bibliography{references}

\begin{thebibliography}{47}
\providecommand{\natexlab}[1]{#1}
\providecommand{\url}[1]{\texttt{#1}}
\expandafter\ifx\csname urlstyle\endcsname\relax
  \providecommand{\doi}[1]{doi: #1}\else
  \providecommand{\doi}{doi: \begingroup \urlstyle{rm}\Url}\fi

\bibitem[Grill et~al.(2020)Grill, Strub, Altch{\'e}, Tallec, Richemond, Buchatskaya, Doersch, Avila~Pires, Guo, Gheshlaghi~Azar, et~al.]{byol}
Jean-Bastien Grill, Florian Strub, Florent Altch{\'e}, Corentin Tallec, Pierre Richemond, Elena Buchatskaya, Carl Doersch, Bernardo Avila~Pires, Zhaohan Guo, Mohammad Gheshlaghi~Azar, et~al.
\newblock Bootstrap your own latent-a new approach to self-supervised learning.
\newblock In \emph{NeurIPS}, 2020.

\bibitem[Chen et~al.(2021)Chen, Xie, and He]{moco_v3}
Xinlei Chen, Saining Xie, and Kaiming He.
\newblock An empirical study of training self-supervised vision transformers.
\newblock In \emph{ICCV}, 2021.

\bibitem[Caron et~al.(2021)Caron, Touvron, Misra, J{\'e}gou, Mairal, Bojanowski, and Joulin]{dino}
Mathilde Caron, Hugo Touvron, Ishan Misra, Herv{\'e} J{\'e}gou, Julien Mairal, Piotr Bojanowski, and Armand Joulin.
\newblock Emerging properties in self-supervised vision transformers.
\newblock In \emph{ICCV}, 2021.

\bibitem[Zhou et~al.(2022)Zhou, Wei, Wang, Shen, Xie, Yuille, and Kong]{ibot}
Jinghao Zhou, Chen Wei, Huiyu Wang, Wei Shen, Cihang Xie, Alan Yuille, and Tao Kong.
\newblock Image {BERT} pre-training with online tokenizer.
\newblock In \emph{ICLR}, 2022.

\bibitem[He et~al.(2022)He, Chen, Xie, Li, Doll{\'a}r, and Girshick]{mae}
Kaiming He, Xinlei Chen, Saining Xie, Yanghao Li, Piotr Doll{\'a}r, and Ross Girshick.
\newblock Masked autoencoders are scalable vision learners.
\newblock In \emph{CVPR}, 2022.

\bibitem[Chen et~al.(2020)Chen, Kornblith, Norouzi, and Hinton]{ssl_simclr}
Ting Chen, Simon Kornblith, Mohammad Norouzi, and Geoffrey Hinton.
\newblock A simple framework for contrastive learning of visual representations.
\newblock In \emph{ICML}, 2020.

\bibitem[He et~al.(2020)He, Fan, Wu, Xie, and Girshick]{moco}
Kaiming He, Haoqi Fan, Yuxin Wu, Saining Xie, and Ross Girshick.
\newblock Momentum contrast for unsupervised visual representation learning.
\newblock In \emph{CVPR}, 2020.

\bibitem[Misra and Maaten(2020)]{ssl_pirl}
Ishan Misra and Laurens van~der Maaten.
\newblock Self-supervised learning of pretext-invariant representations.
\newblock In \emph{CVPR}, 2020.

\bibitem[Wang and Isola(2020)]{understanding_contrastive_ssl}
Tongzhou Wang and Phillip Isola.
\newblock Understanding contrastive representation learning through alignment and uniformity on the hypersphere.
\newblock In \emph{ICML}, 2020.

\bibitem[Chen and He(2021)]{simsiam}
Xinlei Chen and Kaiming He.
\newblock Exploring simple siamese representation learning.
\newblock In \emph{CVPR}, 2021.

\bibitem[Dosovitskiy et~al.(2021)Dosovitskiy, Beyer, Kolesnikov, Weissenborn, Zhai, Unterthiner, Dehghani, Minderer, Heigold, Gelly, et~al.]{vit}
Alexey Dosovitskiy, Lucas Beyer, Alexander Kolesnikov, Dirk Weissenborn, Xiaohua Zhai, Thomas Unterthiner, Mostafa Dehghani, Matthias Minderer, Georg Heigold, Sylvain Gelly, et~al.
\newblock An image is worth 16x16 words: Transformers for image recognition at scale.
\newblock In \emph{ICLR}, 2021.

\bibitem[Li et~al.(2022{\natexlab{a}})Li, Yang, Zhang, Gao, Xiao, Dai, Yuan, and Gao]{esvit}
Chunyuan Li, Jianwei Yang, Pengchuan Zhang, Mei Gao, Bin Xiao, Xiyang Dai, Lu~Yuan, and Jianfeng Gao.
\newblock Efficient self-supervised vision transformers for representation learning.
\newblock In \emph{ICLR}, 2022{\natexlab{a}}.

\bibitem[Govindarajan et~al.(2023)Govindarajan, Sid{\'e}n, Roll, and Lindsten]{dino_vmf}
Hariprasath Govindarajan, Per Sid{\'e}n, Jacob Roll, and Fredrik Lindsten.
\newblock {DINO} as a von mises-fisher mixture model.
\newblock In \emph{ICLR}, 2023.

\bibitem[Oquab et~al.(2024)Oquab, Darcet, Moutakanni, Vo, Szafraniec, Khalidov, Fernandez, HAZIZA, Massa, El-Nouby, Assran, Ballas, Galuba, Howes, Huang, Li, Misra, Rabbat, Sharma, Synnaeve, Xu, Jegou, Mairal, Labatut, Joulin, and Bojanowski]{dinov2}
Maxime Oquab, Timoth{\'e}e Darcet, Th{\'e}o Moutakanni, Huy~V. Vo, Marc Szafraniec, Vasil Khalidov, Pierre Fernandez, Daniel HAZIZA, Francisco Massa, Alaaeldin El-Nouby, Mido Assran, Nicolas Ballas, Wojciech Galuba, Russell Howes, Po-Yao Huang, Shang-Wen Li, Ishan Misra, Michael Rabbat, Vasu Sharma, Gabriel Synnaeve, Hu~Xu, Herve Jegou, Julien Mairal, Patrick Labatut, Armand Joulin, and Piotr Bojanowski.
\newblock {DINO}v2: Learning robust visual features without supervision.
\newblock \emph{TMLR}, 2024.

\bibitem[Assran et~al.(2022)Assran, Caron, Misra, Bojanowski, Bordes, Vincent, Joulin, Rabbat, and Ballas]{msn}
Mahmoud Assran, Mathilde Caron, Ishan Misra, Piotr Bojanowski, Florian Bordes, Pascal Vincent, Armand Joulin, Mike Rabbat, and Nicolas Ballas.
\newblock Masked siamese networks for label-efficient learning.
\newblock In \emph{ECCV}, 2022.

\bibitem[Assran et~al.(2023)Assran, Balestriero, Duval, Bordes, Misra, Bojanowski, Vincent, Rabbat, and Ballas]{pmsn}
Mahmoud Assran, Randall Balestriero, Quentin Duval, Florian Bordes, Ishan Misra, Piotr Bojanowski, Pascal Vincent, Michael Rabbat, and Nicolas Ballas.
\newblock The hidden uniform cluster prior in self-supervised learning.
\newblock In \emph{ICLR}, 2023.

\bibitem[Garrido et~al.(2023)Garrido, Chen, Bardes, Najman, and LeCun]{ssl_duality}
Quentin Garrido, Yubei Chen, Adrien Bardes, Laurent Najman, and Yann LeCun.
\newblock On the duality between contrastive and non-contrastive self-supervised learning.
\newblock In \emph{ICLR}, 2023.

\bibitem[Cuturi(2013)]{sinkhorn_cuturi}
Marco Cuturi.
\newblock Sinkhorn distances: Lightspeed computation of optimal transport.
\newblock In \emph{NeurIPS}, 2013.

\bibitem[Caron et~al.(2020)Caron, Misra, Mairal, Goyal, Bojanowski, and Joulin]{swav}
Mathilde Caron, Ishan Misra, Julien Mairal, Priya Goyal, Piotr Bojanowski, and Armand Joulin.
\newblock Unsupervised learning of visual features by contrasting cluster assignments.
\newblock In \emph{NeurIPS}, 2020.

\bibitem[Kozachenko and Leonenko(1987)]{koleo_estimate}
Lyudmyla~F Kozachenko and Nikolai~N Leonenko.
\newblock Sample estimate of the entropy of a random vector.
\newblock \emph{Problemy Peredachi Informatsii}, 1987.

\bibitem[Beirlant et~al.(1997)Beirlant, Dudewicz, Gy{\"o}rfi, Van~der Meulen, et~al.]{koleo_estimator}
Jan Beirlant, Edward~J Dudewicz, L{\'a}szl{\'o} Gy{\"o}rfi, Edward~C Van~der Meulen, et~al.
\newblock Nonparametric entropy estimation: An overview.
\newblock \emph{International Journal of Mathematical and Statistical Sciences}, 1997.

\bibitem[Sablayrolles et~al.(2019)Sablayrolles, Douze, Schmid, and J{\'e}gou]{spreading_vectors}
Alexandre Sablayrolles, Matthijs Douze, Cordelia Schmid, and Herv{\'e} J{\'e}gou.
\newblock Spreading vectors for similarity search.
\newblock In \emph{ICLR}, 2019.

\bibitem[Caron et~al.(2018)Caron, Bojanowski, Joulin, and Douze]{deep_cluster}
Mathilde Caron, Piotr Bojanowski, Armand Joulin, and Matthijs Douze.
\newblock Deep clustering for unsupervised learning of visual features.
\newblock In \emph{ECCV}, 2018.

\bibitem[Asano et~al.(2020)Asano, Rupprecht, and Vedaldi]{sela}
Yuki~Markus Asano, Christian Rupprecht, and Andrea Vedaldi.
\newblock Self-labelling via simultaneous clustering and representation learning.
\newblock In \emph{ICLR}, 2020.

\bibitem[Caron et~al.(2019)Caron, Bojanowski, Mairal, and Joulin]{ssl_deepercluster}
Mathilde Caron, Piotr Bojanowski, Julien Mairal, and Armand Joulin.
\newblock Unsupervised pre-training of image features on non-curated data.
\newblock In \emph{ICCV}, 2019.

\bibitem[Kukleva et~al.(2023)Kukleva, B{\"o}hle, Schiele, Kuehne, and Rupprecht]{temperature_schedules}
Anna Kukleva, Moritz B{\"o}hle, Bernt Schiele, Hilde Kuehne, and Christian Rupprecht.
\newblock Temperature schedules for self-supervised contrastive methods on long-tail data.
\newblock In \emph{ICLR}, 2023.

\bibitem[Yang et~al.(2022)Yang, Chen, Li, Xie, Lin, and Tao]{etf_neural_collapse}
Yibo Yang, Shixiang Chen, Xiangtai Li, Liang Xie, Zhouchen Lin, and Dacheng Tao.
\newblock Inducing neural collapse in imbalanced learning: Do we really need a learnable classifier at the end of deep neural network?
\newblock \emph{NeurIPS}, 2022.

\bibitem[Fang et~al.(2021)Fang, He, Long, and Su]{minority_collapse}
Cong Fang, Hangfeng He, Qi~Long, and Weijie~J Su.
\newblock Exploring deep neural networks via layer-peeled model: Minority collapse in imbalanced training.
\newblock \emph{Proceedings of the National Academy of Sciences}, 2021.

\bibitem[Deng et~al.(2009)Deng, Dong, Socher, Li, Li, and Fei-Fei]{dataset_imagenet}
Jia Deng, Wei Dong, Richard Socher, Li-Jia Li, Kai Li, and Li~Fei-Fei.
\newblock Imagenet: A large-scale hierarchical image database.
\newblock In \emph{CVPR}, 2009.

\bibitem[Ruan et~al.(2023)Ruan, Singh, Morningstar, Alemi, Ioffe, Fischer, and Dillon]{we_ssl}
Yangjun Ruan, Saurabh Singh, Warren~Richard Morningstar, Alexander~A Alemi, Sergey Ioffe, Ian Fischer, and Joshua~V Dillon.
\newblock Weighted ensemble self-supervised learning.
\newblock In \emph{ICLR}, 2023.

\bibitem[Liu et~al.(2021)Liu, Lin, Cao, Hu, Wei, Zhang, Lin, and Guo]{swin}
Ze~Liu, Yutong Lin, Yue Cao, Han Hu, Yixuan Wei, Zheng Zhang, Stephen Lin, and Baining Guo.
\newblock Swin transformer: Hierarchical vision transformer using shifted windows.
\newblock In \emph{ICCV}, 2021.

\bibitem[Bao et~al.(2022)Bao, Dong, Piao, and Wei]{beit}
Hangbo Bao, Li~Dong, Songhao Piao, and Furu Wei.
\newblock {BE}i{T}: {BERT} pre-training of image transformers.
\newblock In \emph{ICLR}, 2022.

\bibitem[Li et~al.(2022{\natexlab{b}})Li, Andreeto, Ranzato, and Perona]{dataset_caltech101_data}
Fei-Fei Li, Marco Andreeto, Marc’Aurelio Ranzato, and Pietro Perona.
\newblock Caltech 101, 2022{\natexlab{b}}.
\newblock URL \url{https://data.caltech.edu/records/20086}.

\bibitem[Krizhevsky(2009)]{dataset_cifar}
Alex Krizhevsky.
\newblock Learning multiple layers of features from tiny images.
\newblock Technical report, 2009.

\bibitem[Cimpoi et~al.(2014)Cimpoi, Maji, Kokkinos, Mohamed, and Vedaldi]{dataset_dtd}
Mircea Cimpoi, Subhransu Maji, Iasonas Kokkinos, Sammy Mohamed, and Andrea Vedaldi.
\newblock Describing textures in the wild.
\newblock In \emph{CVPR}, 2014.

\bibitem[Nilsback and Zisserman(2008)]{dataset_flowers}
Maria-Elena Nilsback and Andrew Zisserman.
\newblock Automated flower classification over a large number of classes.
\newblock In \emph{2008 Sixth Indian Conference on Computer Vision, Graphics \& Image Processing}, 2008.

\bibitem[Bossard et~al.(2014)Bossard, Guillaumin, and Van~Gool]{dataset_food}
Lukas Bossard, Matthieu Guillaumin, and Luc Van~Gool.
\newblock Food-101 -- mining discriminative components with random forests.
\newblock In \emph{ECCV}, 2014.

\bibitem[Parkhi et~al.(2012)Parkhi, Vedaldi, Zisserman, and Jawahar]{dataset_pets}
Omkar~M Parkhi, Andrea Vedaldi, Andrew Zisserman, and CV~Jawahar.
\newblock Cats and dogs.
\newblock In \emph{CVPR}, 2012.

\bibitem[Xiao et~al.(2010)Xiao, Hays, Ehinger, Oliva, and Torralba]{dataset_sun397}
Jianxiong Xiao, James Hays, Krista~A Ehinger, Aude Oliva, and Antonio Torralba.
\newblock Sun database: Large-scale scene recognition from abbey to zoo.
\newblock In \emph{CVPR}, 2010.

\bibitem[Ericsson et~al.(2021)Ericsson, Gouk, and Hospedales]{ssl_transfer}
Linus Ericsson, Henry Gouk, and Timothy~M Hospedales.
\newblock How well do self-supervised models transfer?
\newblock In \emph{CVPR}, 2021.

\bibitem[Touvron et~al.(2021)Touvron, Cord, Douze, Massa, Sablayrolles, and J{\'e}gou]{deit}
Hugo Touvron, Matthieu Cord, Matthijs Douze, Francisco Massa, Alexandre Sablayrolles, and Herv{\'e} J{\'e}gou.
\newblock Training data-efficient image transformers \& distillation through attention.
\newblock In \emph{ICML}, 2021.

\bibitem[He et~al.(2016)He, Zhang, Ren, and Sun]{resnet}
Kaiming He, Xiangyu Zhang, Shaoqing Ren, and Jian Sun.
\newblock Deep residual learning for image recognition.
\newblock In \emph{CVPR}, 2016.

\bibitem[Zhang et~al.(2021)Zhang, Dai, Yang, Xiao, Yuan, Zhang, and Gao]{vil_model}
Pengchuan Zhang, Xiyang Dai, Jianwei Yang, Bin Xiao, Lu~Yuan, Lei Zhang, and Jianfeng Gao.
\newblock Multi-scale vision longformer: A new vision transformer for high-resolution image encoding.
\newblock In \emph{ICCV}, 2021.

\bibitem[Wu et~al.(2021)Wu, Xiao, Codella, Liu, Dai, Yuan, and Zhang]{cvt_model}
Haiping Wu, Bin Xiao, Noel Codella, Mengchen Liu, Xiyang Dai, Lu~Yuan, and Lei Zhang.
\newblock Cvt: Introducing convolutions to vision transformers.
\newblock In \emph{ICCV}, 2021.

\bibitem[Hendrycks et~al.(2021)Hendrycks, Zhao, Basart, Steinhardt, and Song]{dataset_imagenet_a}
Dan Hendrycks, Kevin Zhao, Steven Basart, Jacob Steinhardt, and Dawn Song.
\newblock Natural adversarial examples.
\newblock \emph{CVPR}, 2021.

\bibitem[Hendrycks and Dietterich(2019)]{dataset_imagenet_c}
Dan Hendrycks and Thomas Dietterich.
\newblock Benchmarking neural network robustness to common corruptions and perturbations.
\newblock \emph{ICLR}, 2019.

\bibitem[Van~der Maaten and Hinton(2008)]{tsne}
Laurens Van~der Maaten and Geoffrey Hinton.
\newblock Visualizing data using t-sne.
\newblock \emph{JMLR}, 2008.

\end{thebibliography}
\clearpage

\appendix

\section{Appendix}

\subsection{Extended related work}
\label{appendix: extended_related_work}

\textbf{Clustering-based self-supervised learning}: Self-supervised learning based on the clustering pretext task is a promising paradigm that has proven to be successful and grown tremendously in recent years. Initial works \citep{deep_cluster, ssl_deepercluster, sela} used a two-stage process of assigning pseudo-labels by clustering the representations and then training the representations using the pseudo-labels as targets. \citet{deep_cluster} proposed uniform pseudo-label sampling that is equivalent to weighting the loss contribution of an input by the inverse of its assigned cluster's size. \citet{swav} used online assignment of pseudo-labels in every batch by clustering the representations over a small window of batches. By adapting this to ViTs, \citet{dino} proposed a self-distillation framework where a teacher network produced the target latent classes. \citet{dino_vmf} demonstrated that this objective corresponds to learning a von Mises-Fisher mixture distribution. \citet{esvit} extended the DINO objective to patch tokens while also leveraging efficient architectures like Swin \citep{swin}. iBOT \citep{ibot} is a recent state-of-the-art method that poses the masked image modeling (MIM) task of BeIT \citep{beit} as a clustering task. Another branch of works have focused on improving the few-shot learning performance of these methods \citep{msn, we_ssl}. Recently, DINOv2 \citep{dinov2} built upon iBOT by making several modifications. By pre-training on the large LVD142M dataset, DINOv2 demonstrated performance surpassing many state-of-the-art visual benchmarks at image and pixel levels. 

\subsection{Sinkhorn-Knopp and probability centering}
\label{appendix: sk_centering_comparison}

Let the batch of $B$ logit scores be denoted as $\bL \in \mathbb{R}^{B \times K}$ with corresponding probability distributions $\bP$. Then, the Sinkhorn-Knopp adjusted probability distributions $\Tilde{\bP}$ are obtained by alternating between normalizing the rows and columns of the matrix $\exp(\bL)$, so that they sum up to $1$. Note that the exponent function is applied element-wise to the matrix. Let the elements of the matrix be denoted as $\bL_{b,k}$. Then, normalizing along the rows yields,
\begin{equation*}
    \Tilde{\bP}_{b, k} \leftarrow \frac{\exp(\bL_{b,k})}{\sum_b \exp(\bL_{b,k})} = \frac{\frac{1}{B}\exp(\bL_{b,k})}{\frac{1}{B} \sum_b \exp(\bL_{b,k})} = \frac{1}{B} \exp(\bL_{b,k} - \log(\frac{1}{B} \sum_b \exp(\bL_{b,k}))).
\end{equation*}
Next, normalizing $\Tilde{\bP}$ along the columns we obtain,
\begin{equation*}
    \Tilde{\bP}_{b, k} \leftarrow \frac{\exp (\bL_{b, k} - \log (\frac{1}{B} \sum_b \exp(\bL_{b, k})))}{\sum_{j=1}^K \exp (\bL_{b, j} - \log (\frac{1}{B} \sum_b \exp(\bL_{b, j})))}.
\end{equation*}
If we consider the initial logit scores $\bL_b$ to be already normalized over the components $K$ such that $\sum_k \exp(\bL_{b, k}) = 1$, then the exponents within the inner sum can be replaced with probabilities. Thus, we obtain the probability distributions after 1 iteration of Sinkhorn-Knopp adjustment as,
\begin{equation*}
    \Tilde{\bP}_{b, k}^{(\mathrm{sk1})} \leftarrow \frac{\exp (\bL_{b, k} - \log (\frac{1}{B} \sum_b \bP_{b,k}))}{\sum_{j=1}^K \exp (\bL_{b, j} - \log (\frac{1}{B} \sum_b \bP_{b,j}))}.
\end{equation*}

On the other hand, the probability centered distributions proposed by \citet{dino_vmf} are obtained as follows, where the centering parameter $c_k$ is calculated as a moving average estimate with momentum parameter $m$:
\begin{align*}
\Tilde{\bP}_{b, k}^{\mathrm{(pc)}} &= \frac{\exp (\bL_{b,k} - c_k)}{\sum_{j=1}^K \exp (\bL_{b,j} - c_j)} ,  &   c_k &\leftarrow mc_k + (1-m) \log \left[ \frac{1}{B} \sum_{b=1}^B \bP_{b,k} \right].
\end{align*}

Comparing the above expressions for $\Tilde{\bP}_{b, k}^{(\mathrm{sk1})}$ and $\Tilde{\bP}_{b, k}^{\mathrm{(pc)}}$ (Eq.~\eqref{eq:sinkhorn_knopp_adjustment} and Eq.~\eqref{eq:centering_adjustment} in section~\ref{sec: mlcd} of the main paper), we observe that probability centering is equivalent to one iteration of Sinkhorn-Knopp with the key distinction that the logit adjustment is calculated as a moving average instead of a batch estimate.

\subsection{KoLeo prototypes implementation}
\label{appendix: koleo_prototypes_implementation}

Given a set of $K$ prototypes $\boldsymbol{W} \in \mathbb{R}^{K \times D}$, to compute the KoLeo estimate of the differential entropy of the prototypes $h_{\mathrm{kl}}(\boldsymbol{W})$, we require computing nearest neighbor distances for each of the prototypes. This can be memory intensive when a large number of prototypes are used. Note that this is not a problem in the case of DINOv2 \citep{dinov2}, as the KoLeo objective is computed between the $B$ data representations in the batch ($B$ is typically much smaller than $K$). Instead, we resort to a stochastic estimate when calculating the loss objective in each batch. For each batch, we randomly partition the prototypes into disjoint partitions containing 2048 prototypes each, $\boldsymbol{W} = \{ \boldsymbol{W}_1, ..., \boldsymbol{W}_T \}, \boldsymbol{W}_t \in \mathbb{R}^{2048 \times D}$. Then, we compute the KoLeo estimate as follows: $h_{\mathrm{kl}}(\boldsymbol{W}) = \sum_{t=1}^T h_{\mathrm{kl}}(\boldsymbol{W}_t)$. This efficient batched implementation adds negligible computational overhead, in terms of both memory and time ($15$ MB additional GPU memory when $K=8192$ and unchanged image throughput).

\subsection{Experimental details}
\label{appendix: experimental_details}

\subsubsection{Hyperparameter settings}
\label{appendix: hyperparameter_settings}

The complete hyperparameter configuration for full-scale iBOT-vMF pre-trainings on ImageNet using ViT-Small/16 and ViT-Base/16 models are provided in Table~\ref{table:app_hyperparams_ibot}. For pre-training on iNaturalist-2018, we use a similar hyperparameter configuration except that we use pre-train both ViT-Small/16 and ViT-Base/16 models for 300 epochs. The complete hyperparameter configurations for MSN and PMSN pre-trainings on the iNaturalist-2018 dataset using the ViT-Small/16 model are provided in Table~\ref{table:app_hyperparams_msn}.

\begin{table}[h]
  \centering
  \scriptsize
  \begin{tabular}{lll}
    \toprule
    Hyper-parameter & ViT-Small/16 & ViT-Base/16 \\
    \midrule
    training epochs & $800$ & $400$ \\
    batch size & $1024$ & $512$ \\
    learning rate & $2\sce{-3}$ & $1.5\sce{-3}$ \\
    warmup epochs & $10$ & $10$ \\
    freeze last layer epochs & $1$ & $3$ \\
    min. learning rate & $1\sce{-6}$ & $2\sce{-6}$ \\
    weight decay & $0.04 \rightarrow 0.4$ & $0.04 \rightarrow 0.4$ \\
    stochastic depth & $0.1$ & $0.1$ \\
    gradient clip & $3.0$ & $0.3$ \\
    optimizer & adamw & adamw \\
    shared head & \cmark & \cmark \\
    fp16 & \cmark & \cmark \\
    \midrule
    momentum & $0.996 \rightarrow 1.0$ & $0.996 \rightarrow 1.0$ \\
    global crops & $2$ & $2$ \\
    global crops scale & $[0.25, 1.0]$ & $[0.32, 1.0]$ \\
    local crops & $10$ & $10$ \\
    local crops scale & $[0.05, 0.25]$ & $[0.05, 0.32]$ \\
    \midrule
    head mlp layers & $3$ & $3$ \\
    head hidden dim. & $2048$ & $2048$ \\
    head bottleneck dim. & $256$ & $256$ \\
    norm last layer & \xmark & \xmark \\
    num. prototypes & $8192$ & $8192$ \\
    vmf normalization & \cmark & \cmark \\
    centering & probability & probability \\
    koleo reg. strength & $0.1$ & $0.1$ \\
    \midrule
    teacher temp. & $0.04 \rightarrow 0.07$ & $0.04 \rightarrow 0.07$ \\
    temp. warmup epochs & $30$ & $50$ \\
    student temp. & $0.1$ & $0.1$ \\
    \midrule
    pred. ratio & $[0.0, 0.3]$ & $[0.0, 0.3]$ \\
    pred. ratio variance & $[0.0, 0.2]$ & $[0.0, 0.2]$ \\
    pred. shape & block & block \\
    \bottomrule
  \end{tabular}
  \caption{Hyperparameter settings for iBOT}
  \label{table:app_hyperparams_ibot}
\end{table}

\begin{table}[h]
  \centering
  \scriptsize
  \begin{tabular}{lll}
    \toprule
    Hyper-parameter & MSN & PMSN \\
    \midrule
    training epochs & $300$ & $300$ \\
    batch size & $1536$ & $1536$ \\
    learning rate & $6\sce{-3}$ & $6\sce{-3}$ \\
    warmup epochs & $15$ & $15$ \\
    min. learning rate & $1\sce{-6}$ & $2\sce{-6}$ \\
    weight decay & $0.04 \rightarrow 0.4$ & $0.04 \rightarrow 0.4$ \\
    stochastic depth & $0.1$ & $0.1$ \\
    gradient clip & $3.0$ & $3.0$ \\
    optimizer & adamw & adamw \\
    fp16 & \xmark & \xmark \\
    \midrule
    momentum & $0.996 \rightarrow 1.0$ & $0.996 \rightarrow 1.0$ \\
    random crops & $1$ & $1$ \\
    local crops & $10$ & $10$ \\
    patch drop rate & $0.15$ & $0.15$ \\
    \midrule
    head mlp layers & $3$ & $3$ \\
    head hidden dim. & $2048$ & $2048$ \\
    head bottleneck dim. & $256$ & $256$ \\
    norm last layer & \cmark & \cmark \\
    num. prototypes & $8142$ & $8142$ \\
    kl penalty weight ($\lambda$) & $1.0$ & $5.0$ \\
    \midrule
    teacher temp. & $0.025$ & $0.025$ \\
    sinkhorn teacher & \cmark & \cmark \\
    temp. warmup epochs & $30$ & $50$ \\
    student temp. & $0.1$ & $0.1$ \\
    \bottomrule
  \end{tabular}
  \caption{Hyperparameter settings for MSN / PMSN}
  \label{table:app_hyperparams_msn}
\end{table}

\subsubsection{MSN and PMSN discusion}
\label{appendix: msn_pmsn_choice}

When pre-training on the iNaturalist-2018 dataset using the ViT-Small/16 model, we run hyperparameter sweeps to select suitable values for the KL penalty strength parameter $\lambda$. We consider the values $\{1.0,5.0,15.0\}$. Based on the linear probing results shown in Table~\ref{table:app_inat18_msn_pmsn_lambda_ablation}, we select $\lambda=1.0$ for MSN and $\lambda=5.0$ for PMSN. Using a higher $\lambda$ with MSN strongly encourages the MLCD to match a uniform prior distribution. When the pre-training dataset is naturally long-tailed, strongly encouraging a uniform prior leads to worse performance. However, we find a smaller penalty strength helps MSN to even outperform PMSN. This indicates that using a weak uniform prior can still be a reasonable choice when pre-training on long-tailed datasets.

\begin{table}[h]
  \centering
  \scriptsize
  \begin{tabular}{lllllll}
    \toprule
    Method & $K$ & $M$ & Overall & Head & Middle & Tail \\
    \midrule
    \multicolumn{5}{l}{\textit{ViT-Small/16}}\\
    MSN ($\lambda = 1$) & 8142 & 3363 & \textbf{43.8} & \textbf{51.4} & \textbf{43.9} & \textbf{41.8} \\
    MSN ($\lambda = 5$) & 8142 & 3123 & 42.3 & 49.6 & 42.5 & 40.4 \\
    MSN ($\lambda = 15$) & 8142 & 1562 & 40.9 & 49.5 & 40.6 & 39.1 \\
    \midrule
    PMSN ($\lambda = 1$) & 8142 & 2919 & 41.4 & 48.9 & 41.3 & \textbf{39.7} \\
    PMSN ($\lambda = 5$) & 8142 & 3005 & \textbf{41.8} & \textbf{50.2} & \textbf{41.9} & \textbf{39.7} \\
    PMSN ($\lambda = 15$) & 8142 & 2927 & 41.0 & 49.1 & 41.4 & 38.7 \\
    \bottomrule
  \end{tabular}
  \caption{iNaturalist-2018 linear probing accuracy with full data}
  \label{table:app_inat18_msn_pmsn_lambda_ablation}
\end{table}

\subsubsection{Transfer linear probing}
\label{appendix:app_lincls_details}

We perform our transfer linear classification experiments on the standard suite of datasets used in self-supervised learning: Caltech101 \citep{dataset_caltech101_data}, CIFAR10, CIFAR100 \citep{dataset_cifar}, DTD \citep{dataset_dtd}, Flowers \citep{dataset_flowers}, Food \citep{dataset_food}, Pets \citep{dataset_pets} and SUN397 \citep{dataset_sun397}. We follow the evaluation protocol from \citet{ssl_transfer} and \citet{ssl_simclr} and train $\normltwo$-regularized linear classifiers. We select the regularization strength among a set of 45 values spaced linearly in the range $[-6, 5]$ in log-space and report the standard evaluation metric for each dataset.

\subsubsection{Sinkhorn-Knopp and mean entropy maximization hyperparameters}
\label{appendix: sk_memax_ablation_settings}

For Sinkhorn-Knopp, we firstly use the vMF normalized version of iBOT and ablate over the number of iterations {1, 3, 5} and find that 3 iterations to work best. This choice for the number of iterations is in agreement with DINOv2 \citep{dinov2}. For both SK (iter=3) and mean entropy maximization and for each compute budget (2, 4 or 8 GPUs for 2 days) we ablate over the following hyperparameters:

\begin{itemize}
    \item vMF normalization: True / False \citep{dino_vmf}
    \item Teacher temperature:
    \begin{itemize}
        \item $\tau = 0.04 \rightarrow 0.07$ (default in \citet{ibot} and  \citet{dino_vmf})
        \item $\tau = 0.05 \rightarrow 0.025$ (default in \citet{we_ssl})
    \end{itemize}
\end{itemize}

For SK(iter=3), we find smaller teacher temperatures to be beneficial as in \citet{we_ssl} and using vMF normalization or not has marginal impact on the performance. For ME-MAX, we find that not using vMF normalization and a smaller teacher temperature leads to better performance. 

\subsubsection{Fine-tuning recipes}
\label{appendix: finetuning_recipes}

\textbf{ImageNet fine-tuning}: We fine-tune on the ImageNet dataset by following the fine-tuning recipe used in BeIT \citep{beit} and iBOT \citep{ibot}, which is found to produce consistently good performance in reasonably fewer epochs compared to other fine-tuning recipes. We fine-tune ViT-Small and ViT-Base models for 200 and 100 epochs respectively and use a batch size of $1024$. We use a layer-wise learning rate decay of $0.75$ for ViT-Small and $0.65$ for ViT-Base. We report the best performance achieved after considering 4 different learning rates: $\{8\sce{-4}, 9\sce{-4}, 1\sce{-3}, 2\sce{-3}\}$.

\textbf{iNaturalist-2018 fine-tuning}: We find the fine-tuning recipe of DeIT \citep{deit} using a smaller learning rate and a larger number of epochs to work better for the iNaturalist-2018 dataset. This is similar to the transfer fine-tuning setup of iBOT \citep{ibot}. We use a fine-tune both ViT-Small and ViT-Base models for 360 epochs using a batch size of $1024$. We use learning rates of $5\sce{-5}$ and $7.5\sce{-6}$ for ViT-Small and ViT-Base respectively.

\subsection{Additional results}
\label{appendix: additional_results}

\subsubsection{Partial prototype collapse in more existing models}
\label{appendix: ppc_extended}

In addition to investigating partial prototype collapse in Table~\ref{table:existing_models_num_unique_prototypes}, we also investigate other self-supervised clustering methods that use a prototypical formulation such as EsViT \citep{esvit} and SWaV \citep{swav}. We demonstrate in Table~\ref{table:existing_models_num_unique_prototypes_extended} that partial prototype collapse also occurs in these methods. We observe that partial prototype collapse also occurs in methods using Resnet50 \citep{resnet}, ViL \citep{vil_model} and CvT \citep{cvt_model} backbones. Though we focus on ViT backbone models in this work, note that partial prototype collapse is not only limited to ViT backbones. 

\begin{table}[h]
  \centering
  \scriptsize
  \begin{tabular}{llll}
    \toprule
    Backbone & Method & \makecell{Initialized\\prototypes\\($K$)} & \makecell{Unique\\prototypes\\($M$)} \\
    \midrule
            ViT-S/16 & DINO & 65536 & 1078 \\
            ViT-B/16 & DINO & 65536 & 804 \\
            ViT-S/16 & DINO-vMF & 65536 & 1157 \\
            ViT-B/16 & DINO-vMF & 65536 & 939 \\
            ViT-S/16 & iBOT & 8192 & 3242 \\
            ViT-B/16 & iBOT-vMF & 8192 & 1170 \\
            ViT-L/16 & iBOT & 8192 & 969 \\
            ViT-L/16 & iBOT$^{**}$ & 8192 & 1037 \\
        Resnet50 & SWaV & 3000 & 1669 \\
        Resnet50 & DINO & 60000 & 984 \\
        Swin-Tiny/W=7 & EsViT & 65536 & 1157 \\
        Swin-Base/W=14 & EsViT & 65536 & 4088 \\
        ViL & EsViT & 65536 & 1741 \\
        CvT & EsViT & 65536 & 1178 \\
    \bottomrule
  \end{tabular}
  \caption{Number of unique prototypes in existing models with $\epsilon=0.025$ (default pre-training: ImageNet-1K, $**$: ImageNet-22K)}
  \label{table:existing_models_num_unique_prototypes_extended}
\end{table}

\subsubsection{Ablation experiment for KoLeo-prototype regularization strength}
\label{appendix: ablation_koleo_regularization}

We conduct an ablation experiment to evaluate the impact of the regularization strength ($\lambda$) of the KoLeo-proto regularization term. We consider a 100 epoch iBOT-vMF pre-training using 8192 prototypes on the Imagenet dataset and evaluate $\lambda = \{ 0.02, 0.1, 0.5 \}$. From Table~\ref{table:ablation_koleo_regularization}, we find that too small $\lambda=0.02$ is unable to fully utilize all the initialized prototypes. We observe improved performance and effective utilization of the prototypes using $\lambda=0.1$ but do not observe further improvements from increasing $\lambda$ further. The main goal of this regularization is to effectively utilize the prototypes. We use the minimum regularization strength $\lambda=0.1$ which is sufficient to achieve this in the experiments in this paper.

\begin{table}[h]
  \centering
  \scriptsize
  \begin{tabular}{llll}
    \toprule
    $\lambda$ & \makecell{Initialized\\prototypes\\($K$)} & \makecell{Unique\\prototypes\\($M$)} & kNN top-1 accuracy \\
    \midrule
        0.0 & 8192 & 1045 & 72.39 \\
        0.02 & 8192 & 4693 & 72.56 \\
        0.1 & 8192 & \textbf{8192} & \textbf{72.62} \\
        0.5 & 8192 & \textbf{8192} & \textbf{72.64} \\
    \bottomrule
  \end{tabular}
  \caption{Ablation experiment for KoLeo-proto regularization strength ($\lambda$)}
  \label{table:ablation_koleo_regularization}
\end{table}

\subsubsection{Computational analysis}
\label{appendix: computational_analysis}

The prototype layer in the self-supervised clustering methods that use a prototypical formulation noticeably contributes to the computational cost of training such methods. The weights associated with $K$ prototypes consists of a $K \times D$ matrix. Typically, the bottleneck dimension $D = 256$. The DINO models use a large $K=65536$ and the prototype layer alone adds an additional 16M trainable parameters to the method. A batch of size $B$, results in the computation of probability distributions of size $B \times K$. For iBOT, which computes the probability distributions for all tokens resulting even larger set of probability distributions of size $B \times T \times K$. The number of prototypes in iBOT is set to $8192$ in the default configuration. Computing such large probability distributions involve heavy memory GPU usage and longer training times. For the default configurations of iBOT with ViT-S/16 backbone, we test the GPU memory use for different numbers of prototypes and batch sizes in the fp16 mode and report the results in Table~\ref{table:computational_cost_prototypes}. Consequently, effective utilization of prototypes can help in reducing the GPU memory required for training such models. For instance, at a batch size of $100$, effectively utilizing only $1024$ prototypes is significantly cheaper ($19.7$ GB GPU memory) than using only $\mathord{\sim}1024$ unique prototypes out of $8192$ initialized prototypes ($28.6$ GB GPU memory). Effective prototype utilization consumes $\mathord{\sim}31 \%$ lower GPU memory compared to the baseline.

\begin{table}[h]
  \centering
  \scriptsize
  \begin{tabular}{lll}
    \toprule
    Batch size ($B$) & Number of prototypes ($K$) & GPU memory (GB) \\
    \midrule
        64 & 1024 & 12.9 \\
        64 & 2048 & 13.7 \\
        64 & 4096 & 15.3 \\
        64 & 8192 & 18.6 \\
        64 & 10240 & 20.2 \\
        100 & 1024 & 19.7 \\
        100 & 2048 & 21.0 \\
        100 & 4096 & 23.5 \\
        100 & 8192 & 28.6 \\
        100 & 10240 & 31.1 \\
    \bottomrule
  \end{tabular}
  \caption{Computational cost of training iBOT method with different number of prototypes}
  \label{table:computational_cost_prototypes}
\end{table}

\subsubsection{ImageNet pre-training with a CNN backbone}
\label{appendix: imagenet_dino_resnet_expt}

In order to explore an additional method and backbone combination, we consider the DINO method pre-training using a Resnet50 backbone. We base our pre-training settings on the hyperparameter configuration in the publicly available DINO codebase \footnote{https://dl.fbaipublicfiles.com/dino/dino\_resnet50\_pretrain/args.txt}. We use the vMF normalized version, use probability centering, 8192 prototypes and train for 100 epochs. In Table~\ref{table:dino_resnet50_expt}, we observe that the KoLeo-proto regularization mitigates the partial prototype collapse and achieves improved performance compared to the baseline and KoLeo-data regularization.

\begin{table}[h]
  \centering
  \scriptsize
  \begin{tabular}{lllll}
    \toprule
    Method & Epochs & $M$ & kNN & Linear \\
    \midrule
        DINO-vMF & 100 & 684 & 59.1 & 69.6 \\
        DINO-vMF (kd) & 100 & 1373 & 59.8 & 70.4 \\
        DINO-vMF (kp) & 100 & 8192 & \textbf{60.1} & \textbf{70.8} \\
    \bottomrule
  \end{tabular}
  \caption{ImageNet classification with full data (kNN, linear) using Resnet50 backbone model}
  \label{table:dino_resnet50_expt}
\end{table}

\subsubsection{Detailed transfer learning results}
\label{appendix: detailed_transfer_learning}

We follow the transfer learning protocol explained in \ref{appendix:app_lincls_details} to evaluate the representations learned by pre-training on Imagenet-1K and iNaturalist-2018 datasets. We report the transfer learning performance in Table~\ref{table:linear_transfer_classification_in1k} and Table~\ref{table:linear_transfer_classification_inat18} for a suite of datasets  on their available validation/test splits. With Imagenet pre-training, we observe that effective utilization of the prototypes using KoLeo-proto regularization produces on par or worse transfer learning performance (based on the overall average) compared to the baseline and KoLeo-data regularization. On the other hand, with iNat-18 pre-training, we observe that effective utilization of the prototypes produces better transfer learning performance (based on the overall average) compared to the baseline and KoLeo-data regularization.

\begin{table*}[h]
  \centering
  \scriptsize
  \begin{tabular}{lllllllllll}
    \toprule
    Method & Cal101 & C10 & C100 & DTD & Flwrs. & Food & Pets & SUN & Avg.  \\
    \midrule
    \multicolumn{9}{l}{\textit{ViT-Base/16}}\\
    DINO-vMF & 94.5 & 97.1 & 86.3 & \textbf{74.8} & \textbf{95.7} & 82.5 & \textbf{94.6} & 68.7 & 86.8 \\
    iBOT-vMF & \textbf{95.5} & \textbf{98.0} & \textbf{88.0} & 74.7 & 94.8 & \textbf{83.6} & 93.9 & \textbf{70.2} & \textbf{87.3} \\
    \rowcolor{white!60!lightgray} iBOT-vMF (kp) & 94.6 & 96.5 & 84.1 & 74.3 & 95.6 & \textbf{83.6} & 94.0 & 69.7 & 86.6 \\ 
    MSN & 92.8 & 96.9 & 85.3 & 73.7 & 92.8 & 80.0 & 93.9 & 66.8 & 85.3 \\
    \midrule
    \multicolumn{9}{l}{\textit{ViT-Small/16}}\\
    DINO-vMF & 93.7 & 96.0 & 83.9 & 74.1 & \textbf{95.0} & 80.1 & 93.9 & 66.6 & 85.4 \\
    iBOT-vMF & 94.1 & \textbf{96.7} & \textbf{84.6} & 72.8 & 94.3 & 80.3 & \textbf{94.1} & 67.3 & 85.5 \\
    \rowcolor{white!60!lightgray} iBOT-vMF (kp) & 94.5 & \textbf{96.7} & 83.9 & 73.7 & 94.4 & \textbf{80.5} & 93.7 & \textbf{67.5} & \textbf{85.6} \\ 
    MSN & 93.1 & 95.9 & 82.9 & 72.0 & 93.3 & 77.8 & 92.8 & 65.5 & 84.1 \\
    WE-SSL & \textbf{94.6} & 93.8 & 81.4 & \textbf{74.9} & 93.9 & 79.1 & 92.8 & 66.5 & 84.6 \\
    \bottomrule
  \end{tabular}
  \caption{Transfer learning classification accuracies when pre-trained on Imagenet-1K and transferred to other datasets}
  \label{table:linear_transfer_classification_in1k}
\end{table*}

\begin{table*}[h]
  \centering
  \scriptsize
  \begin{tabular}{lllllllllll}
    \toprule
    Method & Cal101 & C10 & C100 & DTD & Flwrs. & Food & Pets & SUN & Avg.  \\
    \midrule
    \multicolumn{9}{l}{\textit{ViT-Small/16}}\\
    DINO-vMF & 79.2 & 86.3 & 68.5 & 65.4 & 93.0 & 66.1 & 66.9 & \textbf{48.0} & 71.7 \\ 
    iBOT-vMF & \textbf{80.3} & \textbf{87.0} & 69.2 & 66.3 & 93.6 & 66.4 & 64.2 & 47.4 & 71.8 \\ 
    iBOT-vMF (kd) & 79.3 & 86.3 & 68.1 & 64.7 & \textbf{94.2} & 66.4 & 65.4 & 47.6 & 71.5 \\ 
    \rowcolor{white!60!lightgray} iBOT-vMF (kp) & 80.1 & 86.7 & \textbf{69.9} & \textbf{66.8} & 93.2 & \textbf{66.6} & \textbf{65.5} & 47.4 & \textbf{72.0} \\ 
    MSN ($\lambda = 1$) & 70.8 & 82.3 & 62.8 & 64.0 & 88.6 & 61.5 & 60.3 & 45.3 & 67.0 \\ 
    PMSN ($\lambda = 5$) & 71.2 & 81.3 & 62.6 & 62.1 & 88.4 & 60.7 & 58.1 & 44.1 & 66.1 \\ 
    \midrule
    \multicolumn{9}{l}{\textit{ViT-Base/16}}\\
    iBOT-vMF (kd) & 82.4 & 87.9 & 70.8 & 66.3 & 94.6 & 68.6 & 66.9 & 48.5 & 73.2\\ 
    \rowcolor{white!60!lightgray} iBOT-vMF (kp) & 82.0 & 88.7 & 72.1 & 66.3 & 94.7 & 68.7 & 68.1 & 48.7 & 73.7\\ 
    \bottomrule
  \end{tabular}
  \caption{Transfer learning classification accuracies when pre-trained on iNat-2018 and transferred to other datasets}
  \label{table:linear_transfer_classification_inat18}
\end{table*}

\subsubsection{Representation robustness evaluation}
\label{appendix: robustness_evaluation}

We evaluate the impact of effective prototype utilization on representation robustness by evaluating on standard robustness benchmarks such as Imagenet-A \citep{dataset_imagenet_a} and Imagenet-C \citep{dataset_imagenet_c} datasets. The Imagenet-A dataset contains a set of adversarial images curated from the web, that commonly fool classifiers trained on Imagenet-1K dataset in a supervised manner. Imagenet-C contains images from Imagenet-1K with several types of corruptions and perturbations. We report the robustness metrics in Table~\ref{table:robustness_evaluation}. We observe small but consistent improvement in the robustness to adversarial and corrupted images with effective utilization of prototypes, when compared to the iBOT-vMF baseline. 

\begin{table*}[h]
  \centering
  \scriptsize
  \begin{tabular}{lllll}
    \toprule
    Method & Backbone & Pre-training data & INet-A $\uparrow$ & INet-C $\downarrow$  \\
    \midrule
    DINOv2 $^\ddagger$ & ViT-S/14 & LVD-142M & 33.5 & 54.4 \\
    DINOv2 $^\ddagger$ & ViT-B/14 & LVD-142M & 55.1 & 42.7 \\
    iBOT & ViT-L/16 & INet-22K & 41.5 & 43.9 \\
    DINO & ViT-B/8 & INet-1K & 23.9 & 56.6 \\
    \midrule
    iBOT-vMF & ViT-B/16 & INet-1K & 22.7 & 43.8 \\
    \rowcolor{white!60!lightgray} iBOT-vMF (kp) & ViT-B/16 & INet-1K & $23.5^{(+0.8)}$ & $43.7^{(-0.1)}$ \\
    \midrule
    iBOT-vMF & ViT-S/16 & INet-1K & 14.2 & 51.1 \\
    \rowcolor{white!60!lightgray} iBOT-vMF (kp) & ViT-S/16 & INet-1K & $14.3^{(+0.1)}$ & $50.9^{(-0.2)}$ \\
    \bottomrule
  \end{tabular}\\
   $^\ddagger$: Distilled from a larger ViT-g/14 model pre-trained using DINOv2
  \caption{Representation robustness evaluation on standard robustness benchmarks. We report the accuracy (\%) for Imagenet-A dataset and mean corruption error in \% (lower is better) for Imagenet-C dataset.}
  \label{table:robustness_evaluation}
\end{table*}

\subsection{Visual explanations}
\label{appendix: visual_explanations}

\begin{figure}[t]
    \centering
    \begin{subfigure}{.8\linewidth}
        \centering\includegraphics[width=\linewidth]{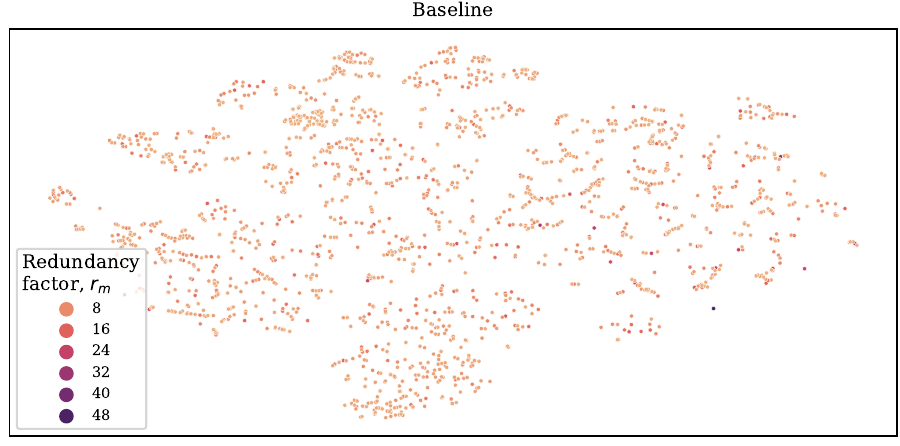}
        \label{fig:tsne_protos_baseline}
    \end{subfigure}
    \hspace{0.04\linewidth}
        \begin{subfigure}{.8\linewidth}
        \centering
        \includegraphics[width=\linewidth]{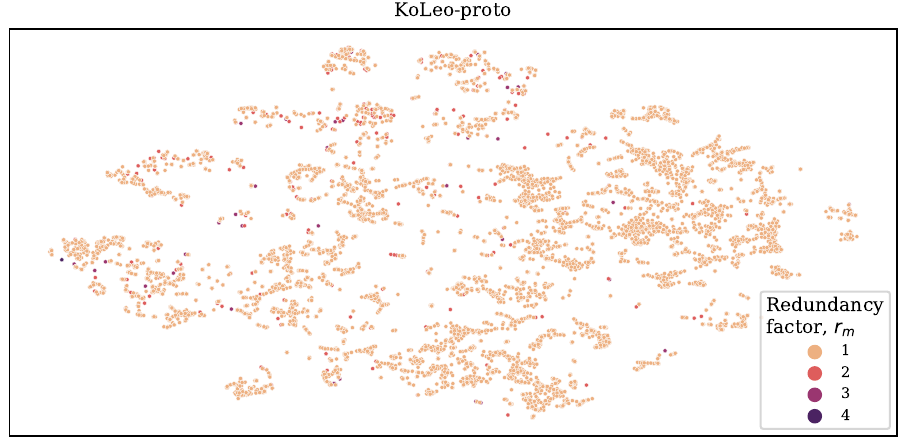}
        \label{fig:tsne_protos_koleo_proto}
    \end{subfigure}
    \caption{t-SNE plot of the $M$ unique prototypes learned by the baseline method and with KoLeo-proto regularization, colored by their redundancy factors $r_m$. There are fewer unique prototypes in the baseline ($M=1806$), noticeable from their sparse spread in the plot. The baseline prototypes are impacted by partial prototype collapse, resulting in high redundancy factors. With KoLeo-proto regularization, the model learns more unique prototypes ($M=7895$) with significantly smaller redundancy factors compared to the baseline. With KoLeo-proto regularization, the method learns diverse prototypes that are well spread over the latent space.}
    \label{fig:tsne_protos}
\end{figure}

\begin{figure}[t]
    \centering
    \begin{subfigure}{.45\linewidth}
        \centering\includegraphics[width=\linewidth]{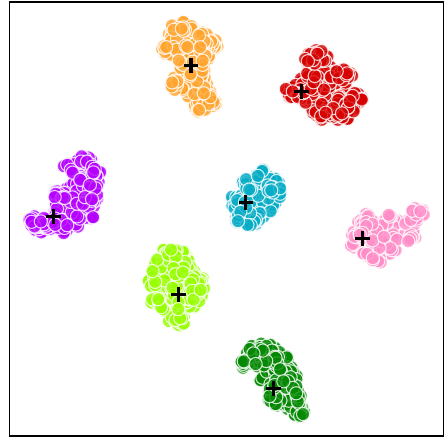}
        \caption{iBOT-vMF baseline}
        \label{fig:tsne_clusters_baseline}
    \end{subfigure}
    \hspace{0.04\linewidth}
    \begin{subfigure}{.45\linewidth}
        \centering
        \includegraphics[width=\linewidth]{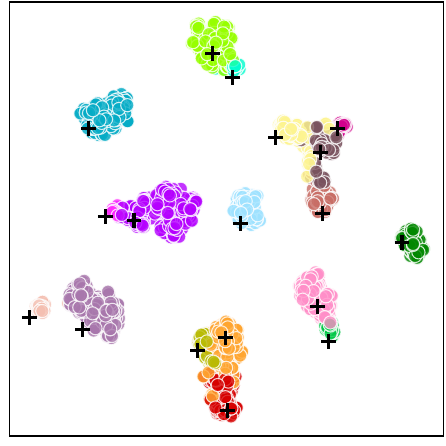}
        \caption{iBOT-vMF with KoLeo-proto}
        \label{fig:tsne_clusters_koleo_proto}
    \end{subfigure}
    \caption{For the exact same set of images, the representations after the head (256 dimensional) are visualized using TSNE plots. The points are colored based on the latent class that they belong to and the corresponding prototypes are denoted using the $+$ marker (the prototype markers are slightly shifted to prevent them from blocking some smaller clusters). The images belong to 7 latent classes in the iBOT-vMF baseline and the same images belong to 18 latent classes when the KoLeo-proto regularization is used. Partial prototype collapse in the baseline results in fewer unique prototypes and coarser clusters. KoLeo-proto regularization encourages diverse prototypes which leads to a more fine-grained clustering of the same data.}
    \label{fig:tsne_clusters}
\end{figure}

\begin{figure}
\setlength\tabcolsep{2pt}
\centering
\begin{tabular}{cc}
 \includegraphics[width=0.45\linewidth]{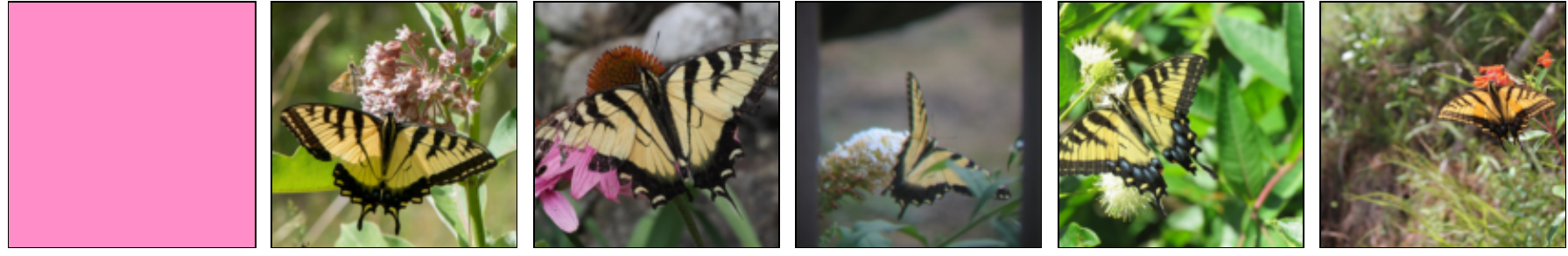} &
 \includegraphics[width=0.45\linewidth]{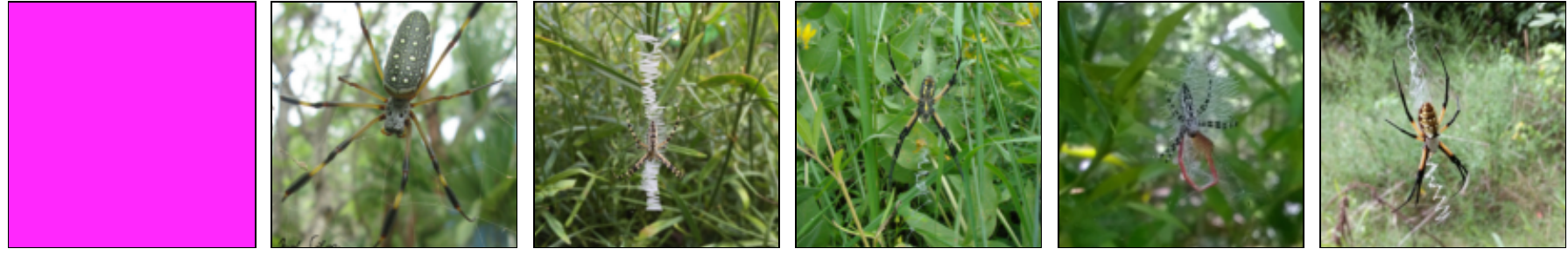} \\
 \includegraphics[width=0.45\linewidth]{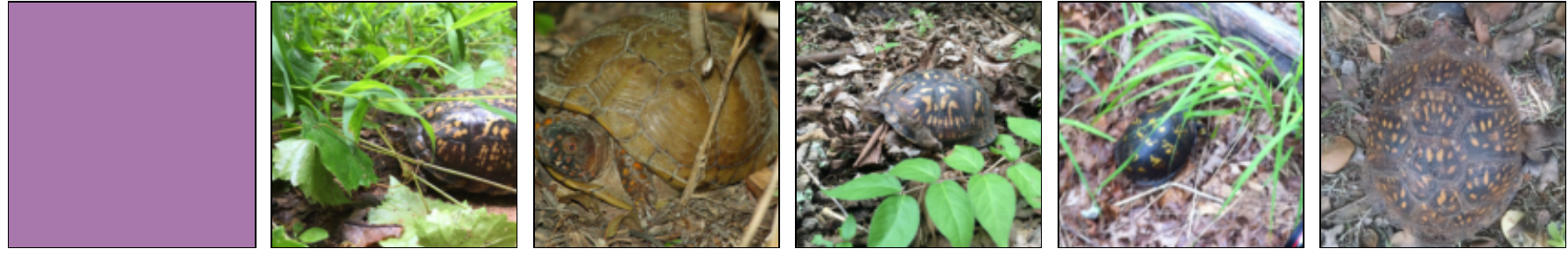} &
 \includegraphics[width=0.45\linewidth]{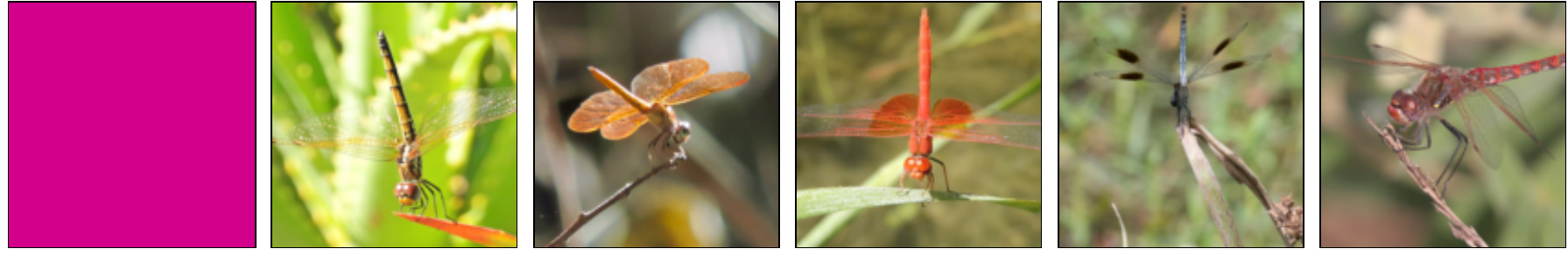} \\
 \includegraphics[width=0.45\linewidth]{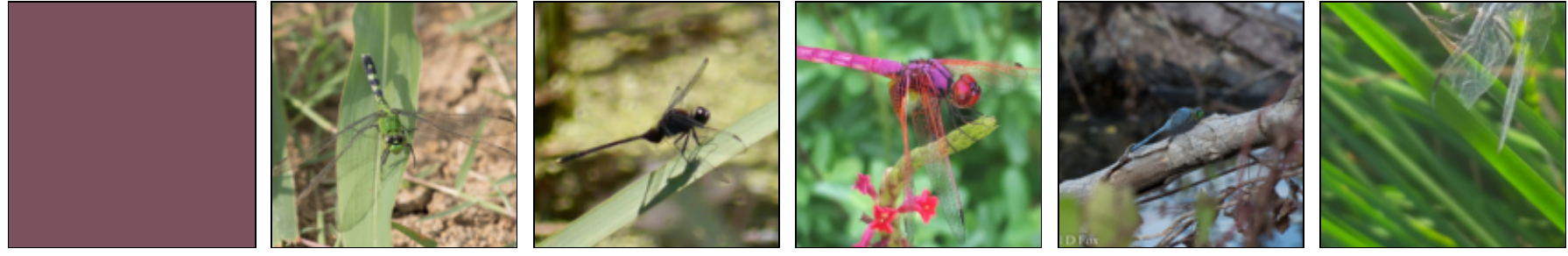} &
 \includegraphics[width=0.45\linewidth]{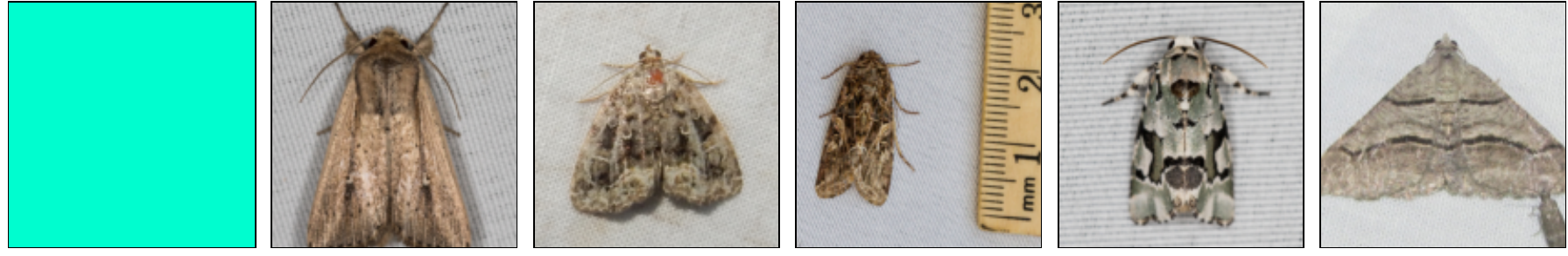} \\
 \includegraphics[width=0.45\linewidth]{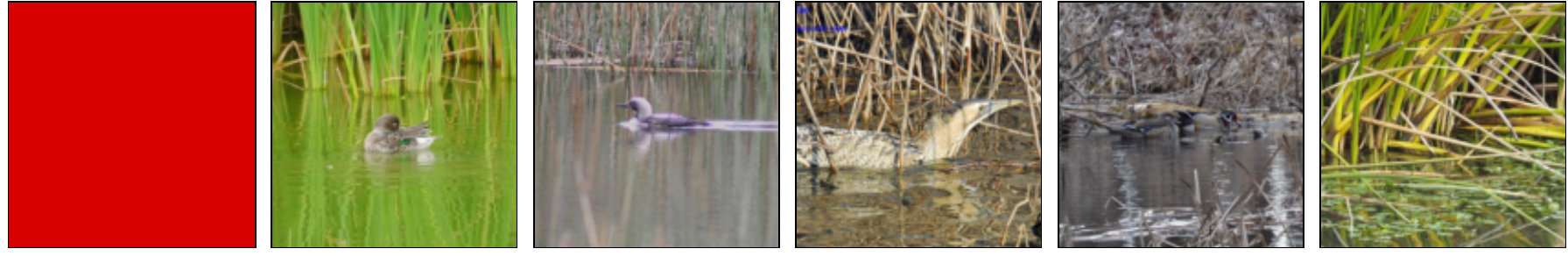} &
 \includegraphics[width=0.45\linewidth]{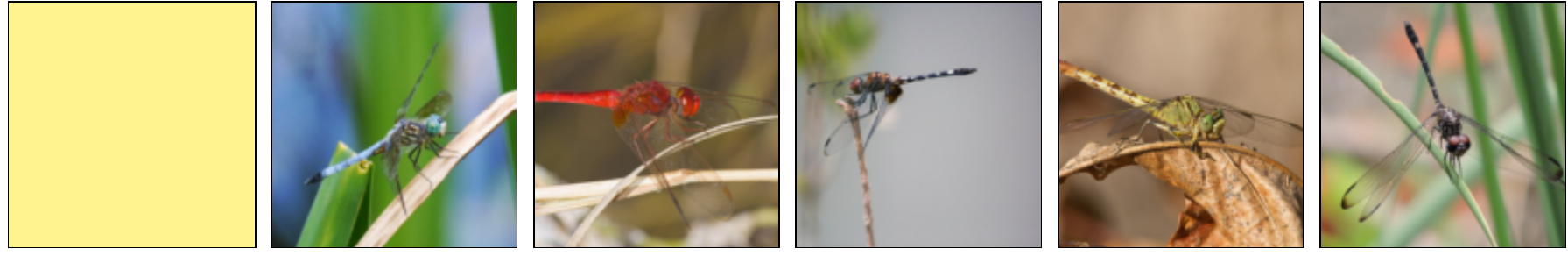} \\
  \includegraphics[width=0.45\linewidth]{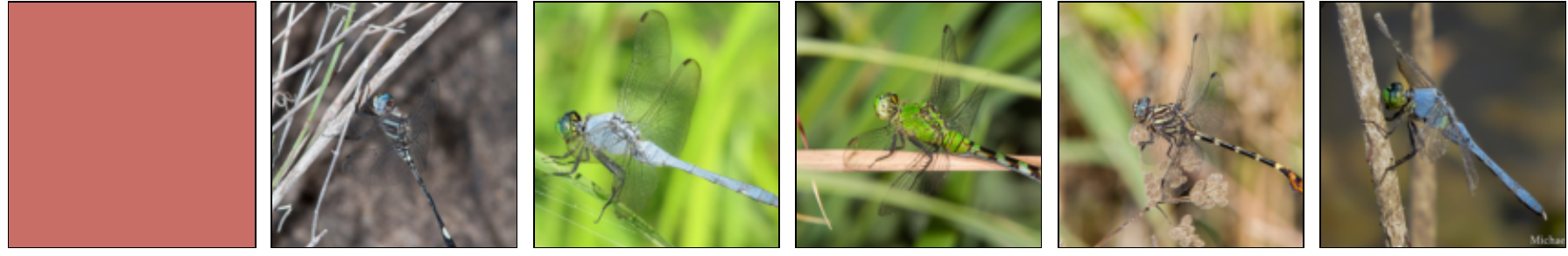} &
 \includegraphics[width=0.45\linewidth]{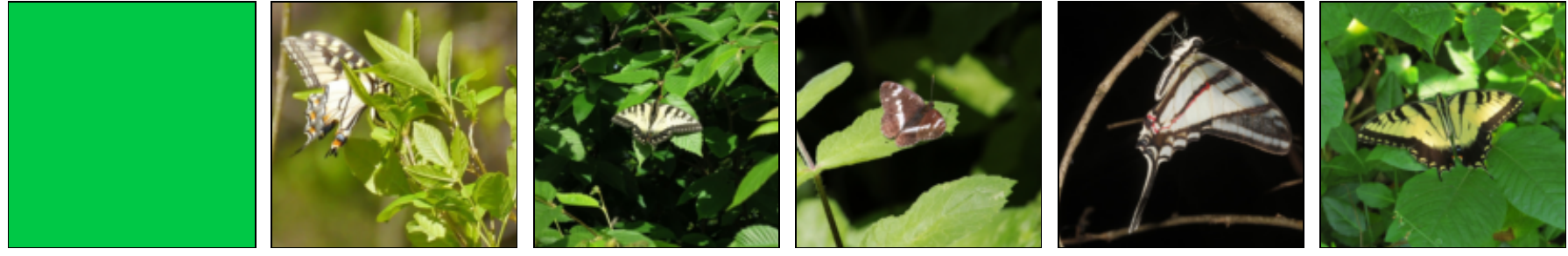} \\
 \includegraphics[width=0.45\linewidth]{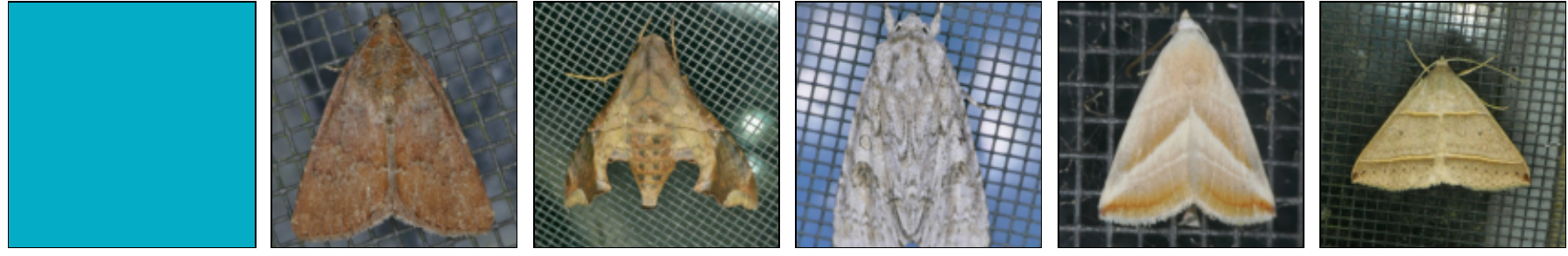} &
 \includegraphics[width=0.45\linewidth]{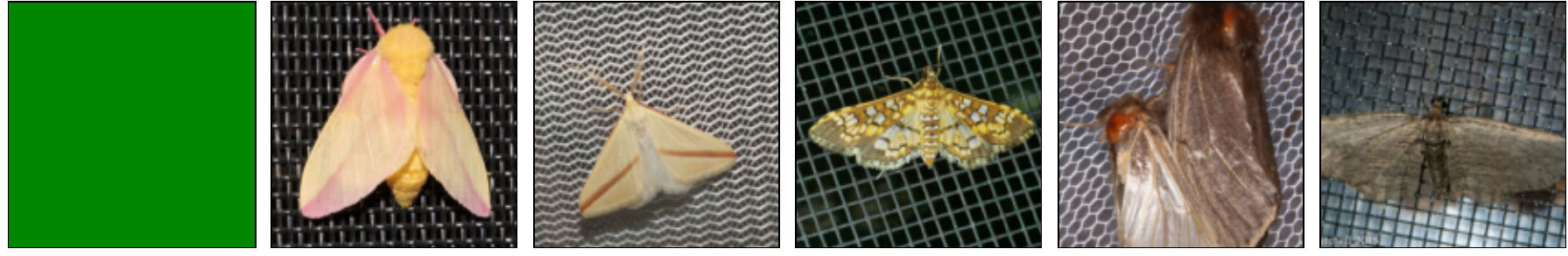} \\
 \includegraphics[width=0.45\linewidth]{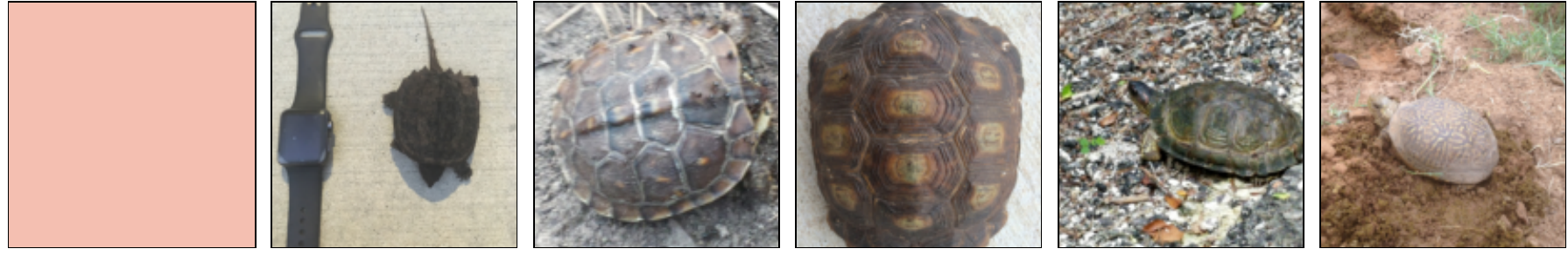} &
 \includegraphics[width=0.45\linewidth]{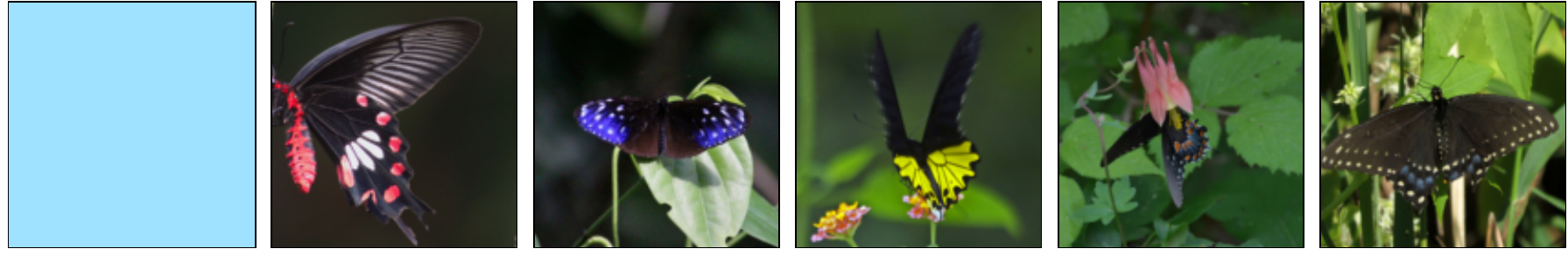} \\
 \includegraphics[width=0.45\linewidth]{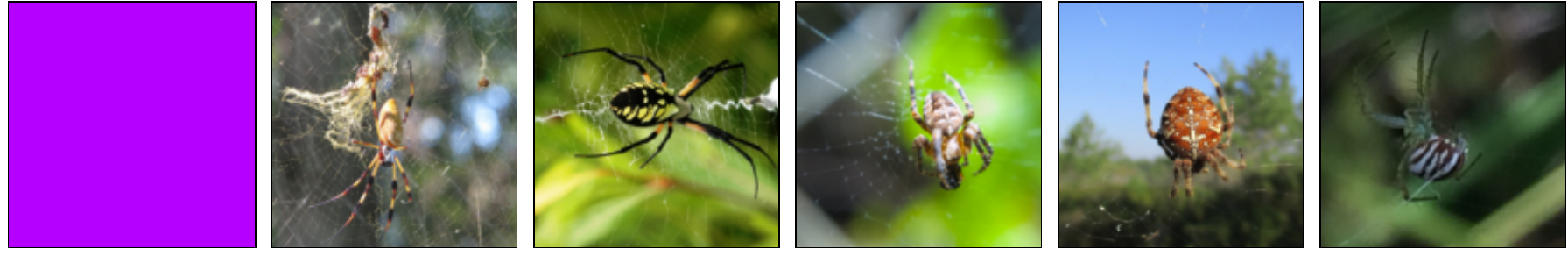} &
 \includegraphics[width=0.45\linewidth]{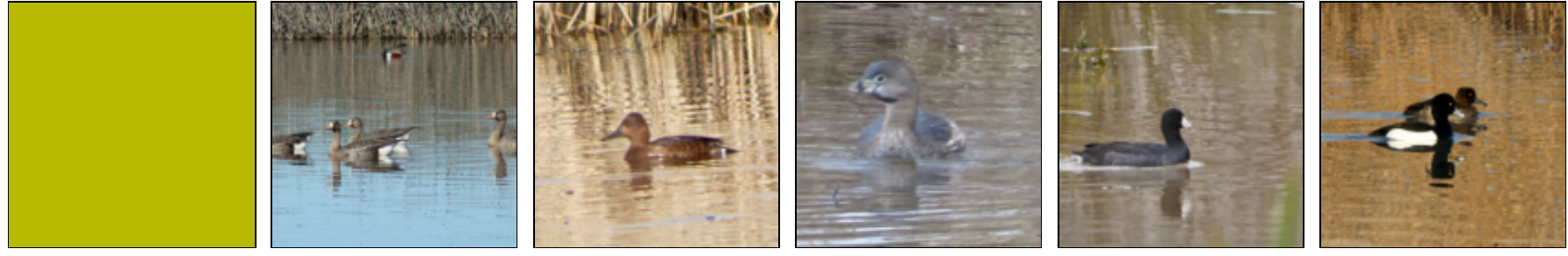} \\
  \includegraphics[width=0.45\linewidth]{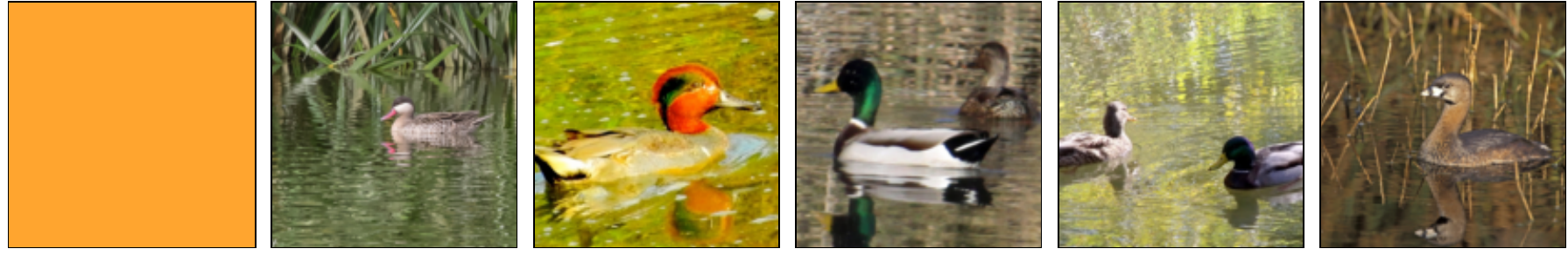} &
 \includegraphics[width=0.45\linewidth]{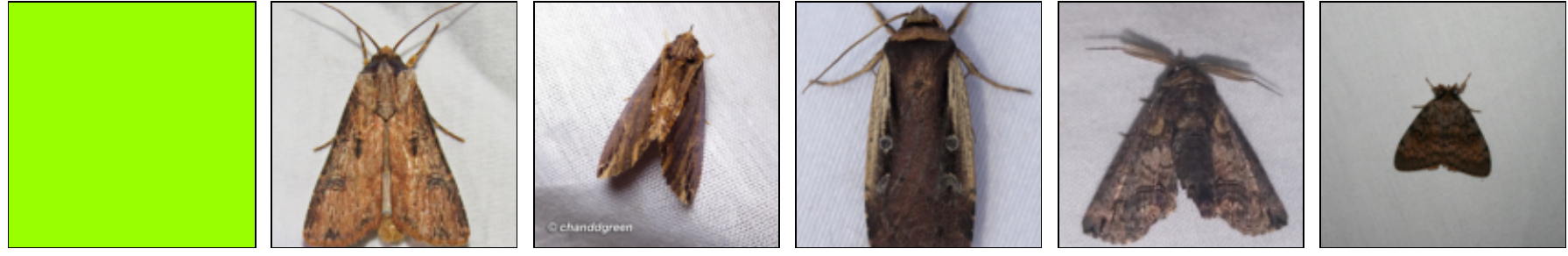} \\
\end{tabular}
\caption{Sample images from the latent classes shown in Figure~\ref{fig:tsne_clusters_koleo_proto} obtained from  iBOT-vMF with KoLeo-proto regularization. Same colors are used to indicate the latent classes. }
\label{fig:example_images_koleo_proto}
\end{figure}

\begin{figure}
\setlength\tabcolsep{2pt}
\centering
\begin{tabular}{cc}
 \includegraphics[width=0.48\linewidth]{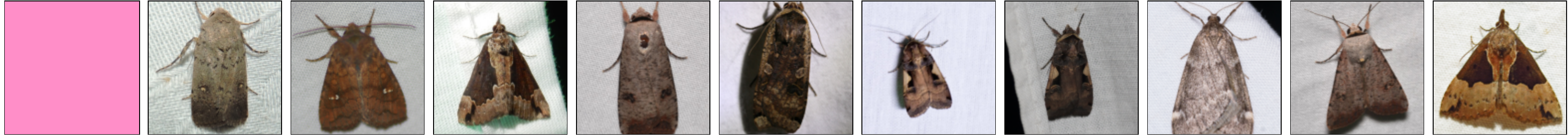} &
 \includegraphics[width=0.48\linewidth]{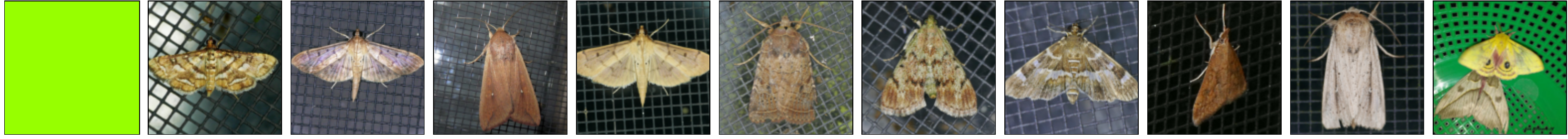} \\
 \includegraphics[width=0.48\linewidth]{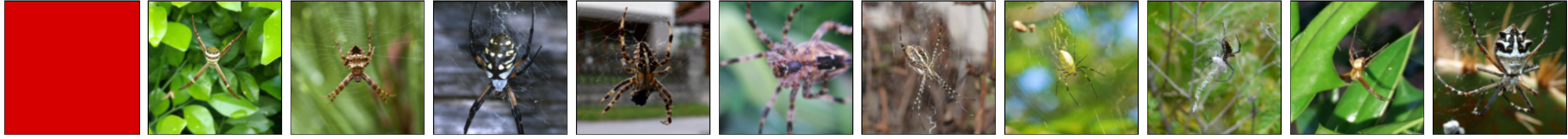} &
 \includegraphics[width=0.48\linewidth]{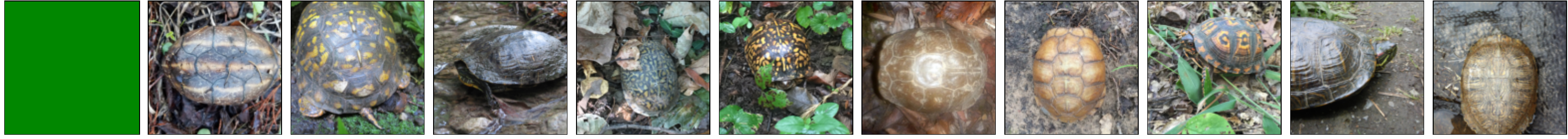} \\
  \includegraphics[width=0.48\linewidth]{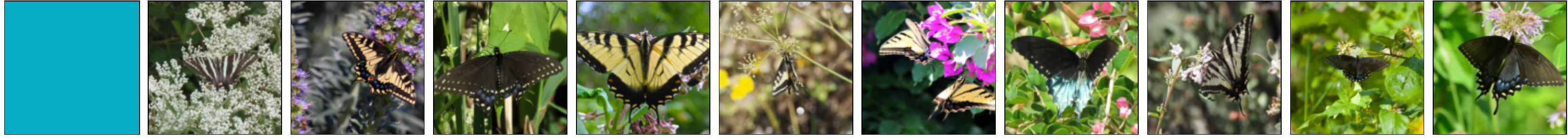} &
 \includegraphics[width=0.48\linewidth]{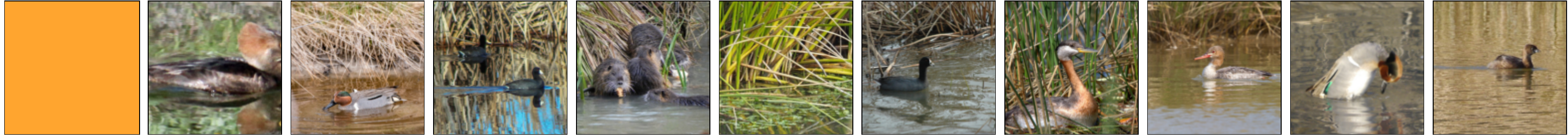} \\
   \includegraphics[width=0.48\linewidth]{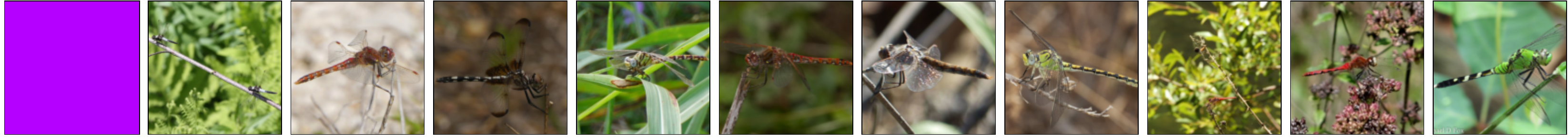} &
  \\
\end{tabular}
\caption{Sample images from the latent classes shown in Figure~\ref{fig:tsne_clusters_baseline} obtained from iBOT-vMF baseline method. Same colors are used to indicate the latent classes.}
\label{fig:example_images_baseline}
\end{figure}

In this section, we present a qualitative comparison of a model trained with and without KoLeo-proto regularization. We compare the iBOT-vMF baseline method based on the ViT-S/16 backbone trained on the iNat18 dataset. We visualize the unique prototypes along with their redundancy factors in Figure~\ref{fig:tsne_protos} using t-SNE plots \citep{tsne}. This illustrates the partial prototype collapse in the baseline and the impact of adding the KoLeo-proto regularization on the prototypes. KoLeo-proto regularization encourages diverse prototypes by spreading them out in the latent space, resulting in a higher number unique prototypes compared to the baseline. We visualize the representations corresponding to images that are assigned to a set of latent classes by the iBOT-vMF baseline in Figure~\ref{fig:tsne_clusters_baseline}. In Figure~\ref{fig:tsne_clusters_koleo_proto}, we visualize the representations corresponding to the exact same images based on the iBOT-vMF model trained with KoLeo-proto regularization. We observe that the KoLeo-proto regularization encourages more fine-grained clusters compared to the baseline. In Figure~\ref{fig:example_images_baseline} and Figure~\ref{fig:example_images_koleo_proto}, we show a few example images belonging to the latent classes shown in Figure~\ref{fig:tsne_clusters}. Without KoLeo-proto regularization, only one coarse latent class is learned containing images of ducks. With KoLeo-proto regularization, this is further divided into three finer latent classes. This demonstrates that the model learns more informative representations which enable it to discriminate between these finer latent classes.

\clearpage

\end{document}